\documentclass[lettersize,journal]{IEEEtran}
\usepackage{amsmath,amsfonts}
\usepackage{algorithmic}
\usepackage{algorithm}
\usepackage{array}
\usepackage[caption=false,font=normalsize,labelfont=sf,textfont=sf]{subfig}
\usepackage{textcomp}
\usepackage{stfloats}
\usepackage{url}
\usepackage{verbatim}
\usepackage{graphicx}
\usepackage{cite}
\usepackage{xcolor}
\usepackage{diagbox}
\usepackage{hyperref}
\usepackage{color}
\newcommand{\red}[1]{\textcolor{red}{#1}}
\usepackage{threeparttable}
\usepackage{upgreek}
\usepackage{booktabs}
\usepackage{multirow}

\begin{document}

\title{GRACE: Designing Generative Face Video Codec via Agile Hardware-Centric Workflow}

\author{Rui~Wan$^\dagger$, Qi~Zheng$^\dagger$, Ruoyu~Zhang, Bu~Chen, Jiaming Liu, Min Li, Minge~Jing,~\IEEEmembership{Member,~IEEE}, Jinjia~Zhou,~\IEEEmembership{Member,~IEEE}, and Yibo~Fan$^\star$,~\IEEEmembership{Member,~IEEE}
\thanks{Rui Wan, Qi Zheng, Ruoyu Zhang, Bu Chen, Min Li, Minge Jing, and Yibo Fan are with the State Key Laboratory of Integrated Chips and Systems, Fudan University, Shanghai 200000, China (e-mail: 20300290034@fudan.edu.cn; qzheng21@m.fudan.edu.cn; ryzhang.beryl@foxmail.com; chenbu@fudan.edu.cn; minli24@m.fudan.edu.cn; mejing@fudan.edu.cn; fanyibo@fudan.edu.cn).}
\thanks{Jiaming Liu is with the Department of Electronic Engineering, Shanghai Jiao Tong University, Shanghai, China. (e-mail: liu\_jiaming@sjtu.edu.cn).}
\thanks{Jinjia Zhou is with Hosei University, Koganei, Tokyo, Japan (e-mail: zhou@hosei.ac.jp).}
\thanks{$^\star$ Corresponding author: Yibo Fan}
\thanks{$^\dagger$ Authors contribute equally.}}

\maketitle

\begin{abstract}
The Animation-based Generative Codec (AGC) is an emerging paradigm for talking-face video compression. However, deploying its intricate decoder on resource and power-constrained edge devices presents challenges due to numerous parameters, the inflexibility to adapt to dynamically evolving algorithms, and the high power consumption induced by extensive computations and data transmission. This paper for the first time proposes a novel field programmable gate arrays (FPGAs)-oriented AGC deployment scheme for edge-computing video services. Initially, we analyze the AGC algorithm and employ network compression methods including post-training static quantization and layer fusion techniques. Subsequently, we design an overlapped accelerator utilizing the co-processor paradigm to perform computations through software-hardware co-design. The hardware processing unit comprises engines such as convolution, grid sampling, upsample, etc. Parallelization optimization strategies like double-buffered pipelines and loop unrolling are employed to fully exploit the resources of FPGA. Ultimately, we establish an AGC FPGA prototype on the PYNQ-Z1 platform using the proposed scheme, achieving \textbf{24.9$\times$} and \textbf{4.1$\times$} higher energy efficiency against commercial Central Processing Unit (CPU) and Graphic Processing Unit (GPU), respectively. Specifically, only \textbf{11.7} microjoules ($\upmu$J) are required for one pixel reconstructed by this FPGA system.
\end{abstract}



\section{Introduction} \label{Introduction}
\IEEEPARstart{I}{n} recent years, there has been a significant increase in the demand for video conferencing and video calls. These videos share a common characteristic where the background remains generally static, requiring focused attention on the speaking individual's facial features. This characteristic embodies strong statistical patterns, leading to the emergence of numerous deep learning-based studies aiming to surpass the compression capabilities of traditional video codecs. Many neural video/image compression methods employ Convolutional Neural Network (CNN) based auto-encoders, whose R-D performance has surpassed that of traditional codecs to some extent\cite{balle2016end,balle2018variational,chen2022exploiting,theis2017lossy,wu2018video,minnen2018joint}. Recently, animation-based generative models have also been applied to talking-face video compression, called AGC (\underline{A}nimation-based \underline{G}enerative \underline{C}odec), leveraging their powerful generative capabilities to significantly reduce bitrate and enhance decoding quality, achieving excellent performance in ultra-low bitrate scenarios\cite{wang2021one, oquab2021low, konuko2021ultra, feng2021generative}. These models, based on prior knowledge, encapsulate face videos into highly compressed elements such as landmarks, keypoints, compressed features, etc., to represent the evolving information of video frames over time. After transmission, these compressed representations can generate complete videos based on a few source frames, a process referred to as image animation.

However, existing learned codecs including AGC primarily emphasize compact representations and the reconstruction ability of generative networks, resulting in complex decoding processes. For example, the decoding time of a learned codec, DVC\cite{lu2019dvc}, is 37.3$\times$ higher than that of X265 (an open-source software for standard video codec)\cite{x265}\cite{SIP-2021-0044}. The learned decoder demands substantial computational resources and energy consumption, thus being constrained to high-performance CPU or GPU platforms, which is absent in edge devices. Edge deployment of AGC decoders needs to solve the following challenges:

\begin{itemize}
\item{\textbf{Extensive number of parameters.} Most edge devices suffer from heavy computation and memory overheads. Taking the decoder of \cite{konuko2021ultra} as an example, it has 45M parameters (\textit{i.e.}, 175 MB in FP32), requiring about 26.72G FLOPS to reconstruct a single frame.}
\item{\textbf{Inflexibility for high-speed evolution model.} The expeditious innovation of algorithms imposes demands on the generality of general-purpose accelerators and the reconfigurability of dedicated accelerators. Faster and low-investment prototyping is particularly needed in research and development environments where quick iteration and experimentation are essential.} 
\item{\textbf{Limited energy consumption.} Power efficiency is critical for edge devices and other power-limited scenarios when massive data needs to be transmitted and processed locally within battery life\cite{10044587}.}
\end{itemize}


Recently, model compression techniques (\textit{e.g.}, quantization\cite{han2015deep}, pruning\cite{han2015learning, han2015deep, han2017ese}, deep compression\cite{han2015deep}, etc.) are commonly applied to minimize the number of operations and the memory storage, but it is crucial to compensate the introduced quality loss. As the most widely used accelerators, GPUs process a large image batch in parallel efficiently at high power consumption\cite{vasudevan2017parallel}, which is challenging for video stream processing on edge devices or embedded systems. Additionally, previous works have targeted application-specific integrated circuits (ASICs) and FPGAs. ASIC achieves optimal performance and energy efficiency through algorithmic fixation and highly customized design, leading to a long development cycle, expensive investment, and high technical risk\cite{chen2014dadiannao, conv2016, han2016eie}. As an alternative, FPGA-based accelerators lower the developing hurdles which provide acceptable throughput at a reasonable price with low power consumption and reconfigurability\cite{qiu2016going, zhang2015optimizing, putnam2014reconfigurable, 8594633, vanderbauwhede2013high, nurvitadhi2017can}. Moreover, High-Level Synthesis (HLS) provides a more efficient and productive way to develop FPGA-based accelerators by allowing designers to describe the design at a higher level of abstraction\cite{canis2011legup,cong2011high,liang2012high}. 
Numerous instances exist wherein CNNs are mapped onto FPGAs to attain accelerated inference\cite{chen2016eyeriss,DBLP:journals/corr/abs-1903-06630, umuroglu2017finn,7544745,7827589,9415618}. Nevertheless, our investigation indicates that prevailing acceleration methodologies predominantly address models tailored for applications such as image classification\cite{cadambi2010programmable,ovtcharov2015accelerating,qiu2016going}, object detection\cite{farabet2009cnp, sankaradas2009massively,cadambi2010programmable,chakradhar2010dynamically,gokhale2014240}, etc. The distinctive nature of AGC decoders poses challenges in direct mapping to conventional accelerator frameworks.

This paper proposes a hardware-software co-design for AGC decoding based on FPGA using HLS. The deployment flow presents an agile and scalable solution for FPGA-based AGC decoders. Models can be compressed using layer fusion and quantization strategies. The co-processing system involves the preliminary establishment of a sparse motion field between two frames based on received key points at first. Subsequently, the reconstruction of the complete video is achieved through the acceleration of the dense motion network and deep generative networks using FPGA. Ultimately, we establish an FPGA prototype for the AGC decoder. The main contributions are summarized as follows:

\begin{itemize}
\item{We undertake a thorough investigation of the AGC algorithm, offering a comprehensive discussion on the novel challenge of its implementation on edge devices. Our analysis provides valuable insights for advancing the future application of high-quality video compression at ultra-low bitrates on edge devices.}
\item{We propose the first prototype system for the AGC decoder deployed on an energy-efficient and configurable edge platform. Layer fusion and post-training static quantization strategies are adopted to perform lightweight optimization of the AGC model, obtaining a 0.25x lighter model. A novel accelerator system is well developed wherein the AGC decoding is implemented on an FPGA platform utilizing the software-hardware co-design.} 
\item{The extensive experimental results indicate that the proposed system achieves higher reconstruction visual quality than traditional codecs at lower than 5 kbps bitrate and attains \textbf{24.9$\times$} and \textbf{4.1$\times$} better energy efficiency compared to CPU and GPU respectively. Only \textbf{11.7} microjoules ($\upmu$J) are required for one pixel reconstructed by this FPGA system. We indeed take a practical step forward towards demonstrating potentials for edge-computing AGC.}
\end{itemize}

The remainder of the paper is organized as follows. In Section \ref{Related works}, we conducted a review of relevant works from two perspectives: video codec algorithms and hardware acceleration of image/video codecs. Section \ref{Dac} details the AGC pipeline with model compression strategies. Section \ref{FPGA} describes the prototype system of the AGC decoder. Section \ref{Exp} shows the various experiment results and system demonstration. We conclude the work in Section \ref{Conclusion}. 

\section{Related works} \label{Related works}

In this section, we provide a comprehensive literature review on the related progress of video compression, which is organized as hybrid video compression, end-to-end video compression, and animation-based generative codec. Additionally, we classified existing FPGA-based CNN accelerators and presented several achievements related to generative image/video codecs.

\subsection{Progress of Video Compression}
\textbf{Hybrid Video Compression.} Cutting-edge video coding standards employ a hybrid framework that consists of prediction, transformation, quantization, and entropy coding. Versatile Video Coding (VVC)~\cite{bross2021overview}, issued by the Joint Video Experts Team (JVET) of the ITU-T Video Coding Experts Group (VCEG) and the ISO/IEC Moving Picture Experts Group (MPEG) in 2020, serves as one of the state-of-the-art video coding standards, significantly improving compression performance of 50\% upon the predecessors HEVC~\cite{sullivan2012overview}. The Alliance for Open Media consortium finalized AV1~\cite{han2021technical} in 2018, which achieves about a 30\% reduction in bit rate compared with its predecessors VP9~\cite{mukherjee2015technical} for the same quality. To fulfill the great demand for Ultra-High Definition (UHD) content, the Audio Video Standard (AVS) in China finalized the first phase of AVS3~\cite{zhang2019recent} in 2019, which outperforms AVS2~\cite{gao2014overview} on bit rate reduction by about 24\%. Recently, a few works have explored the possibility of adopting deep-learning strategies to optimize the compression performance of hybrid video compression by either replacing~\cite{zhao2018enhanced,zhang2018residual,ma2020mfrnet} or optimizing~\cite{li2021deepqtmt,jia2019content,ma2019convolutional,zhao2018enhanced, liu2025frequency} coding tools.

\textbf{End-to-End Video Compression.} Most of the recent end-to-end learning-based image/video compression algorithms~\cite{balle2016end,balle2018variational,chen2022exploiting,theis2017lossy,wu2018video,minnen2018joint, yang2023lossy, wan2025m3, zheng2025unicorn} target to optimize the compression framework based on auto-encoder networks or diffusion models. The first end-to-end image compression framework~\cite{balle2016end} is proposed by Ballé and its successors~\cite{balle2018variational,minnen2018joint} exploit Variational Auto-Encoding (VAE)~\cite{kingma2013auto} with hierarchical priors to achieve superior compression performance than traditional image compression standards such as JPEG~\cite{wallace1991jpeg}, JPEG2000~\cite{skodras2001jpeg2000}, and BPG~\cite{bellard2015bpg}. As for learning-based video compression, Lu et al.~\cite{lu2019dvc} proposed the first end-to-end deep video compression (DVC) framework by implementing all modules of typical video compression with deep neural networks. Hu et al.~\cite{hu2021fvc} proposed a feature-space video compression (FVC) model by performing major operations in feature space. Additionally, Liu et al.~\cite{liu2021deep} also extended to predict and reconstruct videos from the latent vector space through a deep neural network (DNN).

\textbf{Animation-based Video Compression.} With the rapid development of deep generative models such as Generative Adversarial Network (GAN)~\cite{goodfellow2014generative} and VAE, numerous generative compression models based on image animation~\cite{siarohin2019first} have emerged. 
Incorporating the merit of animation, these video compression models can achieve low-bitrate compression by transmitting only the compressed source image and keypoints to reconstruct the whole video through a dense motion estimation network.  
DAC~\cite{konuko2021ultra}, as one of the first generative models exploring animation-based compression, animate the ROI regions of videos with a very low bitrate as well as realistic perceptual quality. Based on DAC~\cite{konuko2021ultra}, the latter HDAC~\cite{konuko2022hybrid} and RDAC~\cite{konuko2023predictive} enhance the reconstructed quality by incorporating frames compressed by HEVC and residual frames compressed by learning-based codec, respectively.
Chen et al.~\cite{chen2022beyond,chen2023compact} model temporal trajectories by compact feature representation instead of key points, and facilitate the synthesis based on a sparse-to-dense estimation strategy. Wang et al.~\cite{wang2023extreme} perform principal component analysis on key points to further explore the potential of motion representations. Instead of key points, Feng et al.~\cite{feng2021generative} and Oquab et al.~\cite{oquab2021low} extract facial landmarks to depict the structure information of face videos, while Wang et al.~\cite{wang2021one} extend to 3D canonical key points and head pose. Considering inaccurate reconstruction due to the single reference frame, several works~\cite{volokitin2022neural,wang2022dynamic,tang2022generative,oquab2022efficient} have been done to leverage multiple reference frames to estimate more comprehensive motion maps.
 
\subsection{FPGA-based CNN Accelerators}

Existing FPGA-based methods for network acceleration can be categorized into two architectures: \textit{overlapping} and \textit{streaming}. The fundamental concept of the overlapping architecture involves designing a reusable, large-scale processing engine instantiated in hardware, with hardware control and data scheduling executed by driver programs running on the CPU\cite{chen2016eyeriss,jia2022fpx,DBLP:journals/corr/abs-1903-06630, electronics12102289}. By properly partitioning the network, the processing engine is invoked multiple times to perform various acceleration tasks. This architecture can adapt flexibly to dynamic models, requiring only modifications to the upper-level driver program without the need for hardware reconfiguration. However, excessive generality may lead to suboptimal acceleration effects, as it fails to fully leverage all computational and bandwidth resources on the hardware platform, resulting in non-linear relationships between acceleration performance and different workloads.

In contrast, the streaming architecture entails designing dedicated hardware modules for each segment of the model, conducting the entire model computation in hardware\cite{umuroglu2017finn,7544745,7827589,9415618}. Through the utilization of pipeline parallelism to enhance throughput by concurrently processing multiple input sets, this architecture allows for specialized optimization of components based on the characteristics of each computation part, thereby achieving optimal acceleration effects. Nevertheless, due to its lack of generality, hardware reconfiguration is required when the model changes to match different computational tasks. In addition, there exist some more distinctive architectures, such as Neurosynaptic processors\cite{steven2016convolutional}.

Numerous studies have already concentrated on the hardware implementation of conventional video codecs, with acceleration targets spanning from H.263\cite{263_1} to H.264\cite{264_1}, HEVC\cite{hevc1, hevc2}, and VVC\cite{vvc1, vvc2, vvc3, vvc4}. However, there exists a paucity of investigations about the hardware acceleration of neural network-based video codecs. 
FPX-NIC\cite{jia2022fpx} stands as the pioneering and exclusive hardware-driven neural video codec architecture designed to facilitate 4K UHD all-intra video coding. It introduces an innovative fully CNN-based auto-encoder in an end-to-end manner, together with a two-stage optimization approach aimed at mitigating the train-test gap through hard quantization. Execution of FPX-NIC is implemented on FPGA devices, manifesting proficient 4K UHD codec capabilities with limited resource utilization. However, due to its intra-frame coding strategy, the inter-frame redundancy characteristics of the videos have not been exploited, leaving the potential for bitrate compression.

Expanding the relevant literature to the domain of image codecs and even virtual image codecs reveals the existence of alternative hardware-based solutions. Shao et al.\cite{electronics12102289} propose an FPGA-based processor suitable for deep learning image compression\cite{balle2016end} to accelerate operations using distinct processing units for layers such as the generalized divisive normalization and upsampling to suit picture encoding and decoding needs. Using the Xilinx Zynq ZCU104 as the hardware implementation platform, the processor throughput reaches 283.4 GOPS at 200 MHz.
VR telepresence is a recently developed technology that can reproduce authentic human presence including real-time expressions in VR environments to reform remote communication. F-CAD\cite{F-CAD} is the first automation tool that supports the whole design flow of hardware acceleration of codec avatar decoders for resource-constrained devices, allowing joint optimization on decoder designs in popular machine learning frameworks and corresponding customized accelerator design with cycle-accurate evaluation. It delivered the highest throughput and efficiency, peaking at 122.1 FPS and 91.6\%.

\section{Hardware-oriented AGC Model} \label{Dac}
\begin{figure*}[t]
\centering
\includegraphics[width=\textwidth]{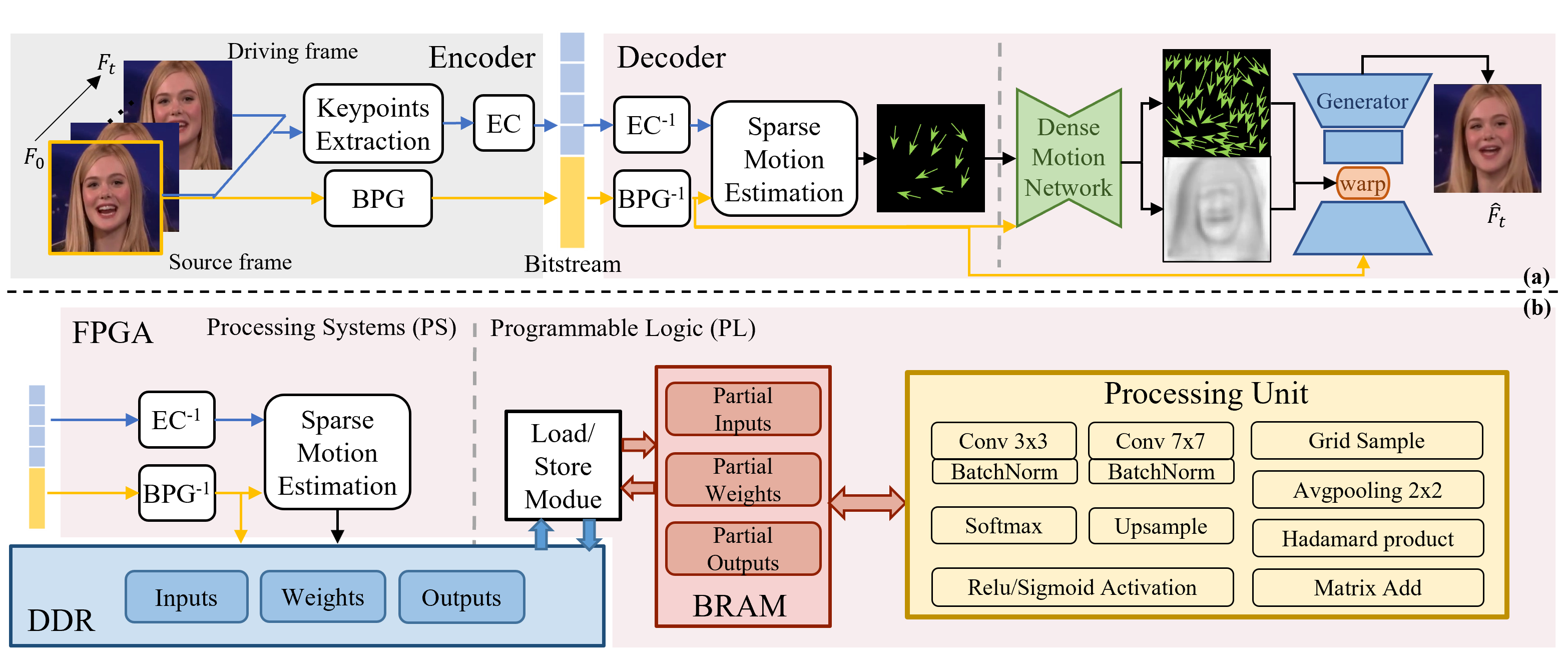}
\caption{The AGC model and decoding system on FPGA platform. \textbf{(a)} is the AGC pipeline. The source frame $F_0$ and extracted keypoints of frames $F_t$ are encoded by BPG and entropy coding, respectively. The decoder estimates the sparse motion between $F_0$ and $F_t$ first. Then the dense motion is predicted by a neural network and $\hat{F_t}$ is reconstructed by a network finally. \textbf{(b)} is the high-level diagram of the FPGA system on which the decoder is deployed. PS and PL serve as preprocessor and neural network accelerator, respectively.}
\label{all}
\end{figure*}

In this section, we present a typical AGC scheme, depicted in Fig.\ref{all}(a). Specifically, the entire codec consists of an encoder with the capability of extracting crucial information and a decoder that employs animation strategies and possesses robust generative capabilities. To map this model onto a hardware platform, we implement hardware-oriented model compression, including post-training static quantization and layer fusion, aiming to alleviate the workload without unduly compromising reconstruction capabilities.

\subsection{AGC Scheme}
\label{sec:baseline}

\subsubsection{Encoder}

The task of the encoder segment is to achieve ultralow bitrate encoding of videos. Our method can be summarized as compressing selected source frames along with motion-related information extracted from video sequences. The segment consists of the following three components:

\textbf{Source frame encoding}. We set the group of pictures (GOP) size to the fixed value of 32 for simplicity. The first frame of each GOP is selected as a source frame, based on which the consecutive frames are reconstructed by the motion estimation and frame generation. Hence the image encoder is necessary to accurately compress the first frame. Following~\cite{konuko2021ultra,konuko2022hybrid}, we compressed the first frame with the BPG image codec\cite{bellard2015bpg} under the given Quantization Parameter (QP), which manifests one of the widely used image compression standards.

\textbf{Keypoint extraction}. We utilize keypoints extracted with the method \cite{siarohin2019animating} to represent the motion information of each frame. Keypoints are predicted with a U-net \cite{ronneberger2015u} architecture neural network in a self-supervised fashion. In the last layer of the decoder, one heatmap is generated as a detection confidence map for each keypoint. Each keypoint coordinate is estimated using average operation. With proper methods for training, the keypoints extracted by the network are semantically consistent in most cases, which enables image animation based on keypoint trajectories.

\textbf{Bitstream generating}. Each keypoint is stored as a 2-dimension vector with FP16 precision and compressed using Entropy Coding (EC) methods. The coded information and compressed source images form the bitstream which is sent to the decoder together.

\subsubsection{Decoder}
\begin{figure}[t]
\centering
\includegraphics[width=0.49\textwidth]{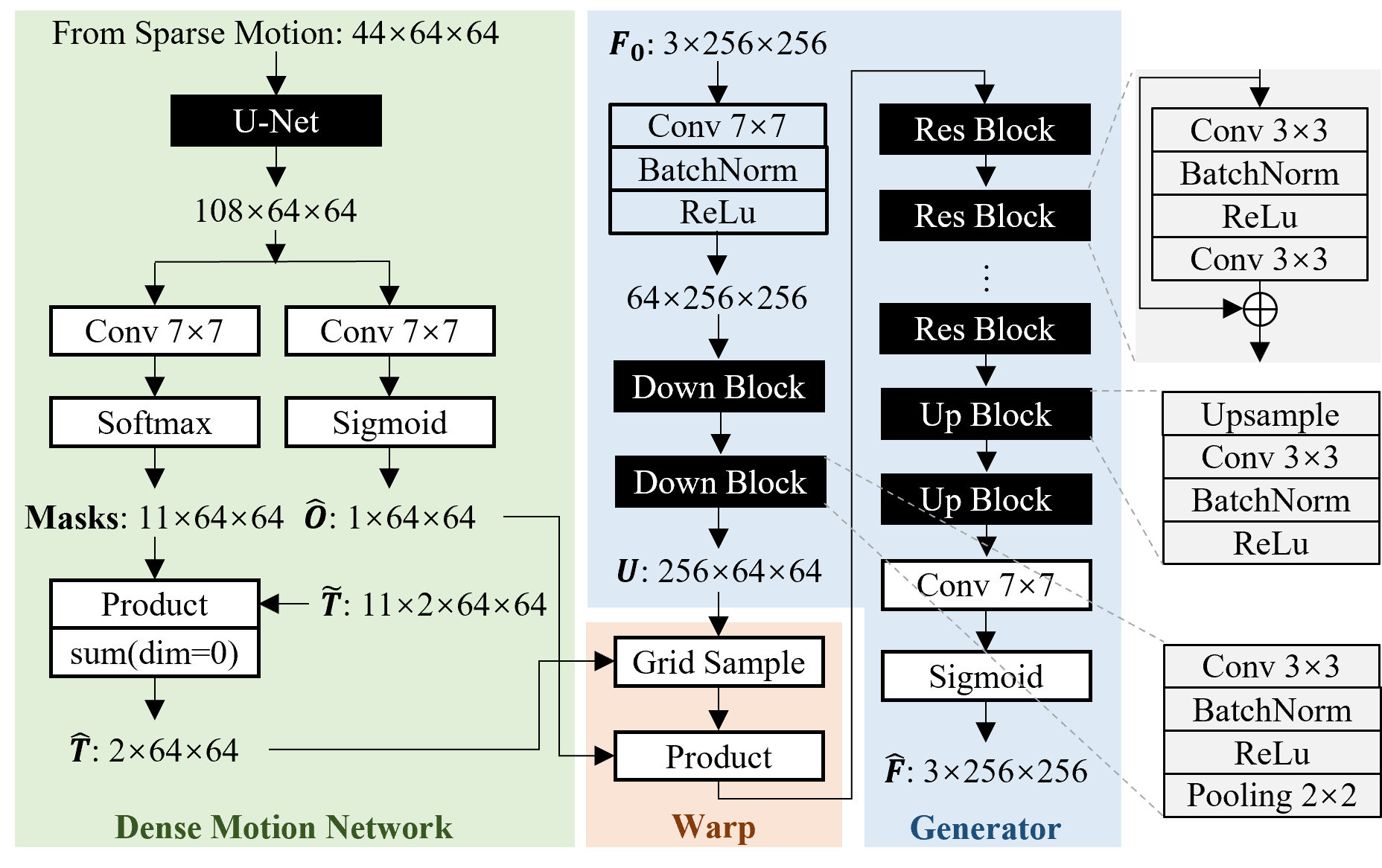}
\caption{The detailed architecture of Dense Motion Network and frame generation. Here, we take the example of an input frame resolution of 256$\times$256 and a keypoint count of 10. The background color here corresponds one-to-one with the modules' color in Fig.\ref{all}(a). }
\label{PL}
\end{figure} 

The main task of our decoder segment is to reconstruct frames from the input bitstream. To reconstruct the $t^{th}$ non-intra driving frame of the current GOP (denoted as $F_t$), we need to utilize the decoded source frame $F_0$ and 2-dimension keypoints $K_t$,$K_0$  extracted from the two frames. The decoder segment consists of the following four components: 

\textbf{Bitstream parsing}. The source frames and keypoints are parsed from the bitstream before further processing.

\textbf{Sparse motion estimation}. To reconstruct $F_t$ (non-intra driving frame) from $F_0$ (source frame), a precise backward optical flow map $\hat{T}_{F0 \leftarrow Ft}$ should be computed. Since the motion patterns in $\hat{T}_{F0 \leftarrow Ft}$ are aligned with $F_t$ rather than $F_0$, leading to a misalignment issue, the optical map cannot be directly computed. Instead, we conduct a preliminary estimation of the backward optical flow (denoted as $\tilde{T}_{F0 \leftarrow Ft}$, comprising $K$ flow maps) :
\begin{equation}
    \label{taylor_approximation}
    \begin{aligned}
    \begin{array}{c}
    \tilde{T}_{F0 \leftarrow Ft}(z,k) = z + K_0(k) – K_t(k),\\
    k=1,2,…,K.
    \end{array}
    \end{aligned}
\end{equation}
where $ K_0(k) $, $ K_t(k) $ respectively denotes $k^{th}$ keypoint coordinates of the driving frame  $ K_0 $ and driving frame $ K_t $, while $z$ denotes arbitrary pixel in the driving frame. For each keypoint, we apply the equation to warp the source frame and obtain a deformed frame roughly depicting motion in the neighborhood of the keypoint. Additionally, $K$ heatmaps roughly indicating where each warping transformation happens are computed in the following manner:
\begin{equation}
    \label{heatmap}
    \begin{aligned}
    \begin{array}{c}
    H_k(z) = exp(\frac{\| K_t(k) – z \|_2^2}{\sigma}) - exp(\frac{\| K_0(k) – z \|_2^2}{\sigma}),\\
    k=1,2,…,K.
    \end{array}
    \end{aligned}
\end{equation}
The source frame $F_0$, the $K$ transformed frames $F_0^1$,$F_0^2$,…, $F_0^K$ and heatmaps $H_1$,$H_2$,…,$H_K$ are used as input of the next module. 

\textbf{Optical flow alignment and occlusion map generation by Dense Motion Network}. Frames generated from the last module are concatenated and processed by a U-net, as shown in Fig.\ref{PL}. Two parallel convolutional layers (with softmax activation) are introduced following the output layer of U-net. The first one is utilized for generating $K+1$ masks $A_0,A_1,…,A_K$ to indicate where each warping transformation holds. We conduct the following computation to obtain the precise backward optical flow $\hat{T}_{F0 \leftarrow Ft}$:
\begin{equation}
    \label{precise_optical_flow}
    \hat{T}_{F0 \leftarrow Ft}(z) = A_{0}(z)z + \sum_{i=1}^{K}{A_{i}(z) \tilde{T}_{F0 \leftarrow Ft}(z,i)}
\end{equation}
The second convolutional layer is utilized for computing an occlusion map $\hat{O}_{F0 \leftarrow Ft}$. The occlusion map indicates which parts of the driving image can be reconstructed by warping the source image and where should be inpainted.

\textbf{Frame generation}. As shown in Fig.\ref{PL}, the module comprises a convolutional network $Generator$ to restore a non-intra driving frame $F_t$ with corresponding source frame $F_0$, backward optical flow $\hat{T}_{F0 \leftarrow Ft}$ and occlusion map $\hat{O}_{F0 \leftarrow Ft}$. The input frame first goes through two downsampling blocks, after which we obtain a feature map $U$. $U$ is warped with $\hat{T}_{F0 \leftarrow Ft}$. Hadamard product of the deformed feature map $U’$ and $\hat{O}_{F0 \leftarrow Ft}$ is computed consequently and sent to the decoder part of $Generator$. Finally, a high-quality reconstructed frame is generated at the end of the decoder. 
\\

\subsubsection{Training Method}

The codec pipeline is trained on a large training set of videos depicting talking faces. Reconstruction perceptual loss proposed in \cite{johnson2016perceptual} is selected for optimization, which allows the keypoint predictor network to extract keypoints that enable our codec to precisely reconstruct each frame. An equivariance loss is utilized to enforce the keypoint extraction network to be equivariant to random geometric transformations \cite{siarohin2019animating}, which significantly improves the stability of keypoint extraction. A GAN loss proposed in \cite{mao2017least} is implemented at the end of our codec pipeline to enhance reconstruction quality.

\subsection{Model Compression}
\label{sec:integerAGC}
Deploying CNN models on resource-constrained embedded or edge devices necessitates lightweight optimizations to conserve space and reduce computational overhead. Quantization and operator fusion are essential measures. When CNN computations encounter memory bottlenecks, redundant and inefficient computational resources become a critical concern, making data movement a pivotal issue for performance optimization.

\subsubsection{Layer Fusion}

The BatchNorm layer is commonplace following the convolution layer as it often reduces training time and enhances generalization. It normalizes intermediate tensors, ensuring zero mean and unit variance per channel, as expressed by the following,

\begin{equation}
\label{BN}
\hat{x} = \gamma \frac{x - \mu}{\sqrt{\sigma^2 + \epsilon}} + \beta.
\end{equation}

\noindent $\mu$ and $\sigma^2$ are the mean and variance of a batch, and $\epsilon$ is a small constant to avoid zero division. Convolutions can be simplified as,
\begin{equation}
    \label{conv_eq}
    y = wx + b.
\end{equation}
\noindent Convolution layer and BatcnNorm layer can be merged into a new convolution-like pattern by calculating\cite{markuvs2018fusing,repair}:
\begin{subequations}\label{fusion}
\begin{align}
\hat{y} = w'x + b',\\
w' = \frac{\gamma}{\sqrt{\sigma^2 + \epsilon}}w,\\
b' = \frac{\gamma(b-\mu)}{\sqrt{\sigma^2+\epsilon}}+\beta.
\end{align}
\end{subequations}
\noindent As a result, with appropriate rescaling and bias-shifting, the new Conv+BatchNorm layer reduces the frequency of data transfers and saves space for intermediate data, remaining functionally equivalent to the original two layers at the same time.

\subsubsection{Quantization}
Quantization is a powerful tool for performing both computations and memory accesses with lower precision data, which enables a significant reduction in model size, memory width, and inference time. A floating point model can be quantized to $N_p$-bit precision after two quantizer parameters are derived, the scale $Z$ and the zero-point $Z$. The mapping between the floating point values and the integer values is simplified as follows:
\begin{equation}
    w_{q} = round(\frac{w_f}{S}) + Z,
\end{equation}
\begin{equation}
    w_{dq} = S(w_{q} - Z).
\end{equation}
where $w_{q}$ is quantized interger weight, $w_f$ is floating point weight, and $w_{dq}$ is the weight after inverse quantization. In our framework, the GPU-trained AGC model with floating point 32 precision is quantized into an int 8-bit model without fraction bits. A per-channel quantization technique is adopted to maintain higher accuracy wherein weights of each output channel in a convolution layer or linear layer are independently quantized.

The typical dynamic quantization involves mapping weights into int8 representation, and dynamically converting activations to int8 on the fly before performing computation between weights and activations. However, these activations are read and written to the memory in the format of floating point, resulting in the improvement room of latency. Hence, we conduct Post-Training Static Quantization (PTSQ) in our framework to enable both integer computation and integer memory accesses. In static quantization, the calibration dataset is required to feed into the network for recording and computing the data distributions of different activations, which is done by observer modules that are inserted in the network ahead of time. These distributions are used to derive specific quantizer parameters for different activations at inference time. Hence quantized integer values can be consistently passed between operations without repetitive mapping between floats and ints, bringing about a remarkable speed-up compared to dynamic quantization.

\section{Accelerator System} \label{FPGA}

In this section, a novel accelerator system in which the AGC decoding is implemented on an FPGA platform using software-hardware co-design. In particular, a universal development flow is proposed to establish connectivity between the trained model and the deployment platform. Considering the general purpose and flexibility, the FPGA system has been meticulously designed as an overlapping accelerator architecture. Constrained by the computational and storage resources of the FPGA, various optimizations have been applied to enhance the performance of the accelerator.

\subsection{Development Flow}

\begin{figure}[!t]
\centering
\includegraphics[width=0.5\textwidth]{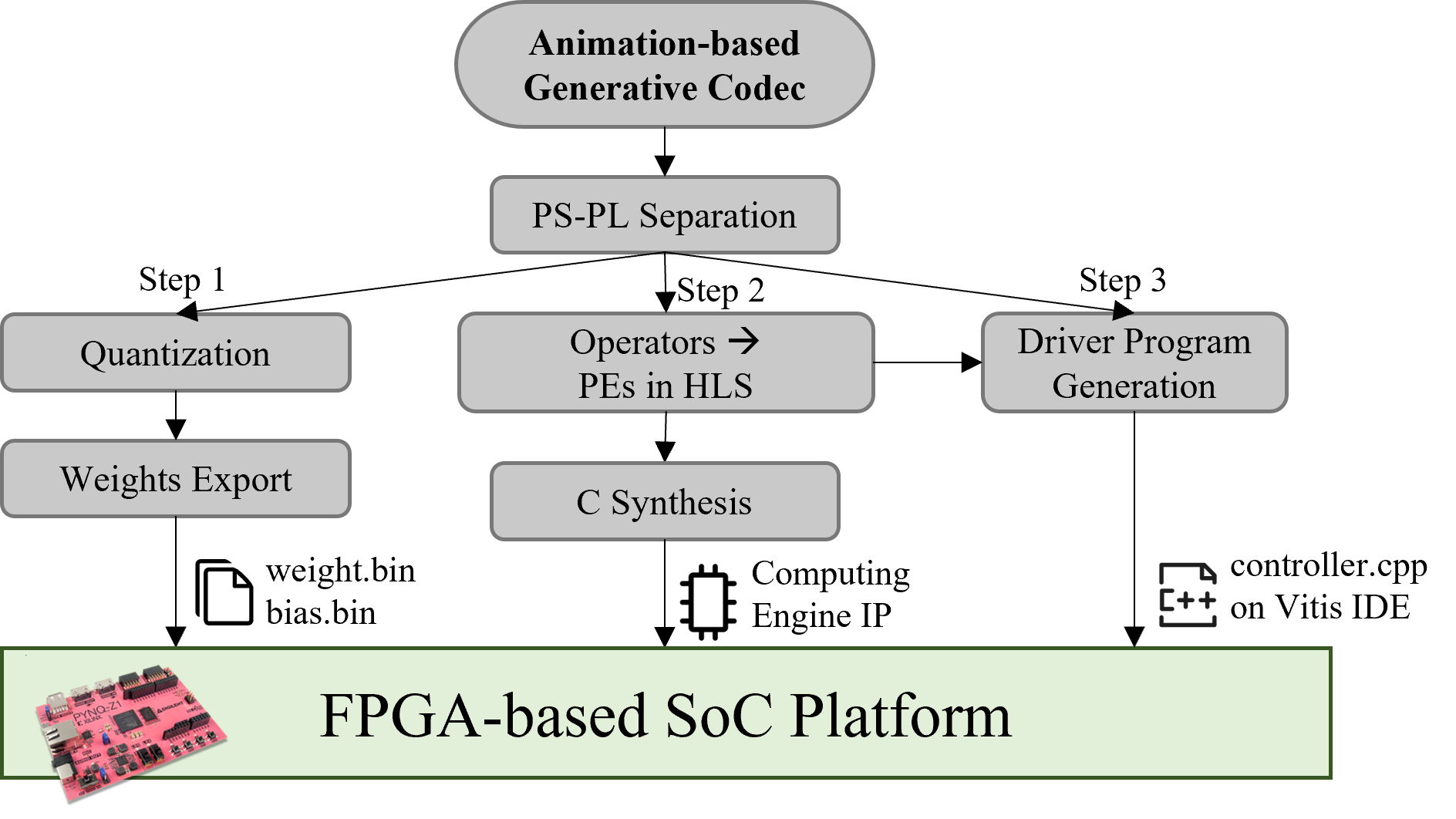}
\caption{The overall deployment flow of AGC.}
\label{flow}
\end{figure}

Fig.\ref{flow} delineates the overall deployment workflow of AGC decoding onto the FPGA board. Initially, a rigorous analysis of the original codec methodology is conducted, delineating the entire process into two distinct components to be implemented in the Processing System (PS) and Programmable Logic (PL) domains, as indicated by the vertical dashed line in Fig.\ref{all}. The PS is conducive to the execution of flexible functionalities requiring frequent development and maintenance, whereas the PL is suitable for handling complex computational tasks. Additionally, considering the finite resources of the FPGA, tasks with lower workloads tend to be delegated to the PS for execution, thereby freeing up computational resources, storage space, and bandwidth of PL for more intricate hardware tasks. In AGC applications, the PS domain encompasses data preprocessing tasks, including entropy decoding, BPG decoding, and sparse motion estimation, while the PL domain accommodates dense computations (as shown in Fig.\ref{PL}) amenable to acceleration by specialized engines. 

Subsequently, a post-training static quantization process is applied to the computations designated for the PL domain, resulting in the derivation of compressed weight and bias. The data are subsequently exported and stored in the off-chip memory on the FPGA board. Furthermore, the operators earmarked for acceleration in the PL domain undergo HLS design, with C synthesis exporting them as Register Transfer Level (RTL) representations of accelerator IP. Concurrently, a driver program, authored in a high-level language (\textit{e.g.} C++), is requisite to control the activation of individual Processing Engines (PEs) within the accelerator. 

Ultimately, the compressed weight data, accelerator IP, and driver program are individually deployed onto the off-chip memory, programmable hardware resources, and ARM processor within the FPGA-based System-on-Chip (SoC). The deployment flow outlined above bridges the research gap between software-based AGCs and hardware-based accelerators for them, offering a universal and adaptable FPGA deployment process and a comprehensive array of selectable PEs for prospective AGC algorithms. 

\subsection{High-level Diagram of FPGA System}

\begin{figure}[h]
\centering
\includegraphics[width=0.4\textwidth]{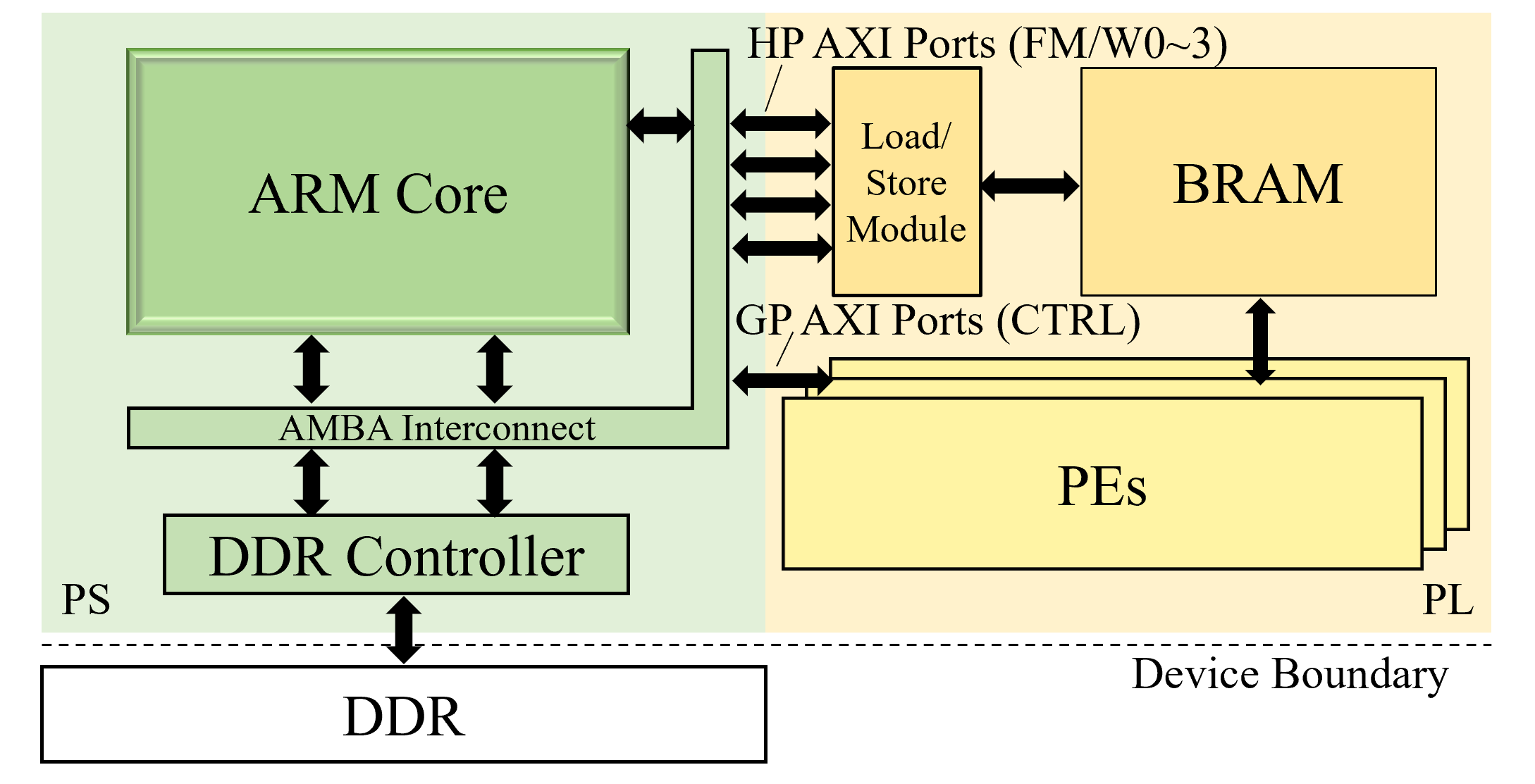}
\caption{SoC-level diagram. The SoC is divided into two distinct subsystems: the Processing System (PS) and the Programmable Logic (PL). }
\label{SOC}
\end{figure}

Fig.\ref{all}(b) presents a high-level schematic of the entire system and Fig.\ref{SOC} shows a SoC-level diagram. Specifically, the entire FPGA system comprises two main components: PS and PL. The PS consists of numerous components, including the Application Processing Unit (APU), comprising two ARM Cortex-A9 processors, the Advanced Microcontroller Bus Architecture (AMBA) interconnect, DDR3 memory controller, and various peripheral controllers. The ARM CPU in the PS serves as a preprocessor, utilizing a BPG decoding program to decode source frames from the bitstream. Additionally, it employs entropy decoding to extract keypoints for each frame from the bitstream. The decoded information, including the source frame and keypoints for both the current driving frame and the source frame, enables the estimation of sparse motion between the two frames. Considering data exchange between the FPGA board and external platforms, the PS is also responsible for read/write operations to portable memory storage devices such as SD cards.

The Double Data Rate Synchronous Dynamic Random-Access Memory (DDR) memory, as an off-chip storage resource, stores all the operational data, including sparse motion, the source frame, network weights, and network results. Managed by the DDR controller in the PS, DDR acts as an intermediary hub facilitating substantial data exchanges between the PS and PL. 

The PL constitutes a meticulously designed accelerator IP, serving as a peripheral connected to the interconnect, incorporating Block Random Access Memory (BRAM) as on-chip caches and a Processing Unit (PU) consisting of multiple Processing Engines (PEs). The BRAM stores partial data for on-chip processing. Operators supported by PU is illustrated in Fig.\ref{all}(b). In addition to conventional CNN operators (such as Conv2d, Activation, Pooling, etc.), indispensable specialized operators for AGC decoding (including grid\_sample, upsampling, and Hadamard product) are also incorporated. Following Vivado HLS\footnote{https://docs.xilinx.com/r/en-US/ug1399-vitis-hls/Introduction}, the PL IP is synthesized with interfaces using the appropriate protocol (\textit{i.e.}, \texttt{m\_axi} for a pointer to I/O data and \texttt{s\_axilite} for scalar passing) to communicate with the external. Considering the substantial volume of input-output feature maps (\texttt{FM}) and weight parameter data (\texttt{W}), as well as the stringent latency requirements, high-performance AXI interfaces (AXI HP Ports) are employed to connect with the PS. Conversely, parameters such as kernel size, stride, and presence of bias (bundled as \texttt{CTRL}) demand less bandwidth, thus establishing connectivity with the PS through general-purpose AXI interfaces (AXI GP Ports) suffices. A driver program running in the PS enables modules and one of the PEs within the PL according to the specified network structure, achieved through manipulating functional registers of the PL. Simultaneously, the operational status of the PL is relayed back to the PS through status registers.

\subsection{Tile-based Processing Dataflow}

Due to constraints in on-chip memory, it is impractical to cache all data in on-chip RAMs. Therefore, a strategy of tile-wise computation is employed to partition the entire computational task. Taking the convolution operator as an example, the data flow within the accelerator will be elucidated.

\begin{figure}[!ht]
\centering
\includegraphics[width=0.48\textwidth]{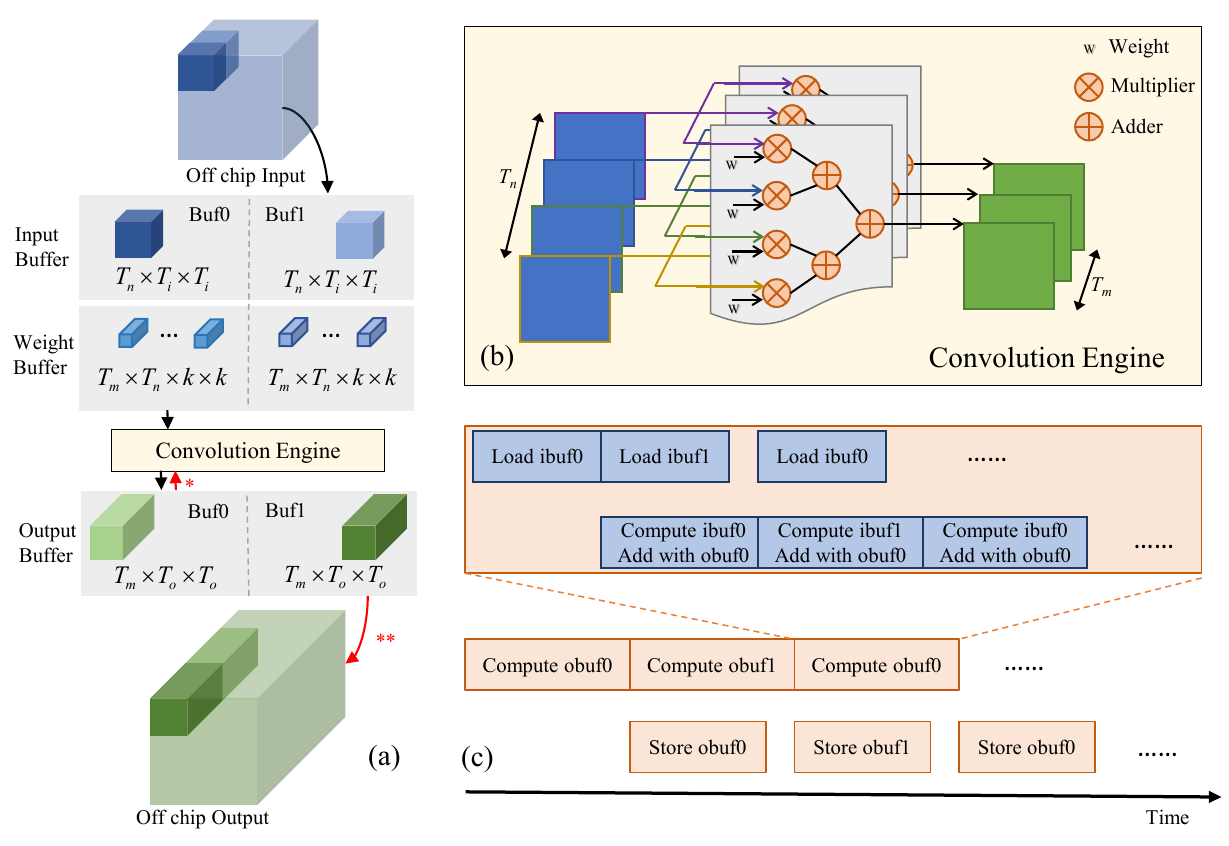}
\caption{An example of off-on-off-chip tile-based convolution dataflow. Data movement marked with ``\red{*}'' keeps happening when the output tile stores partial sums. Data movement marked with ``\red{**}'' only happens when all the tiles along channel\_in dimension are computed.}
\label{dataflow}
\end{figure} 

The data flow within the convolution operator accelerator is depicted in Fig. \ref{dataflow}. Input feature maps, with shape $H_{in}\times W_{in}\times C_{in}$, and convolution kernels, with shape $k\times k\times C_{in}\times C_{out}$, are pre-stored in off-chip DDR. Initially, input tiles of size $T_i\times T_i\times T_n$ and kernels of size $k\times k\times T_n\times T_m$ are loaded into on-chip buffers. Subsequently, convolution computations are performed on the PE, yielding intermediate results of size $T_o\times T_o\times T_m$, which are temporarily stored in the output buffer. New input tiles and kernels are continuously loaded onto the chip, where the PE conducts convolution operations on the fresh input, accumulating the results with the intermediate values stored in the output buffer. This process is iterated $C_{in}/T_n$ times until the tile in the output buffer represents the final accumulated result, which is then transmitted back to the off-chip DDR. This entire process is repeated until all output tiles are obtained, forming the complete output feature map of size $H_{out}\times W_{out}\times C_{out}$.

Bias and activation are implemented within the process. In the case of bias, initialization of the output buffer with bias occurs before the beginning of computations for each output tile. Consequently, when PE undertakes the first accumulation of convolution operation results with the data in the output buffer, the bias is naturally incorporated into the outcome. If activation is needed, subsequent to the completion of computations for each output tile, activation operations are performed on each data before storage to DDR.

\subsection{Design for Parallelism}

\begin{figure}[t]
\centering
\includegraphics[width=0.49\textwidth]{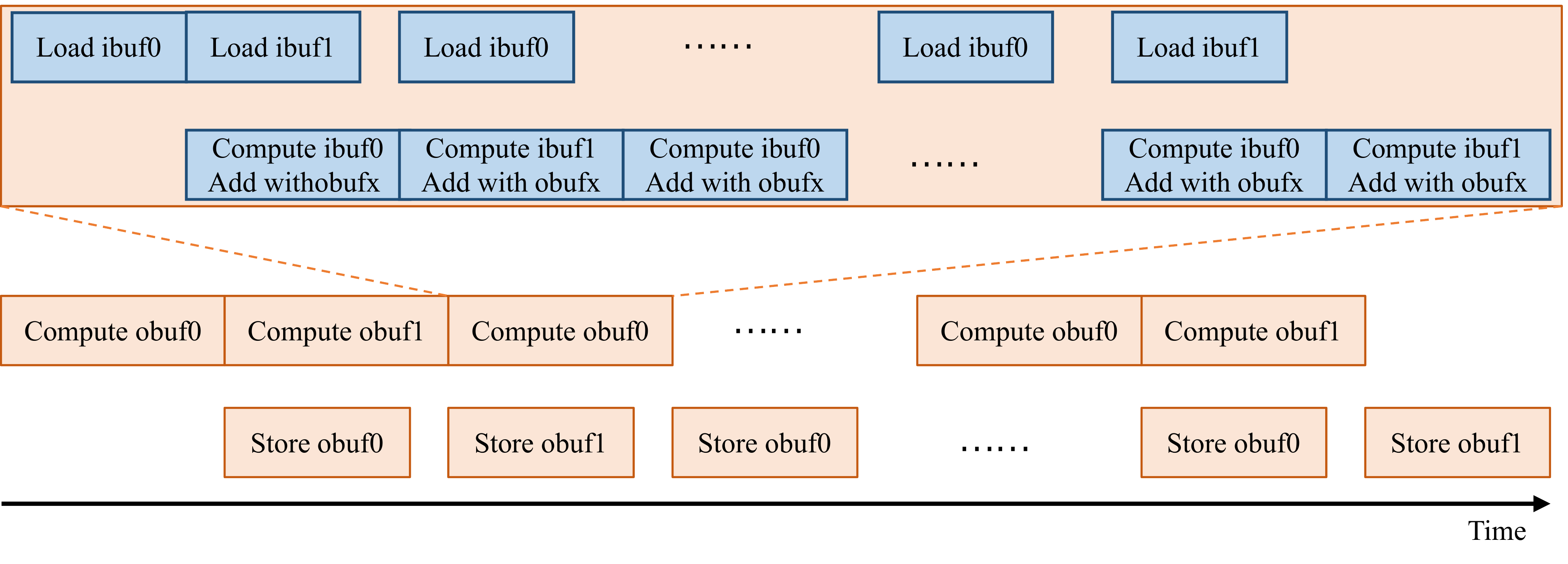}
\caption{LOAD/STORE and computation overlap due to ping-pong buffers, so as to hide data transmission delay.}
\label{pingpong}
\end{figure} 

To fully exploit the system's acceleration capabilities, it is essential to leverage all the resources provided by the FPGA hardware platform, \textit{i.e.}, enabling the concurrent processing of multiple tasks. Computation and communication represent the two primary constraints for optimizing system throughput, where the system may be either computation-bound or memory-bound\cite{zhang2015optimizing}.

\textbf{System optimization.} Pipelining a loop allows the operations of the loop to be implemented in a concurrent manner. Fig.\ref{pingpong} illustrates a pipeline processing architecture comprised of dual buffering. This architecture facilitates the overlap of computation and data transfer times, thereby reducing the total time for the entire process to be determined by the longer of the two sub-processings. Dual buffering is employed for both input feature buffers and output feature buffers, as evident in Fig.\ref{dataflow}. The total time for each output tile to obtain the final result is determined by the longest duration among the load times of all input tiles, convolution computations, and the store time for that specific output tile.

\begin{figure}[!t]
\centering
\includegraphics[width=0.3\textwidth]{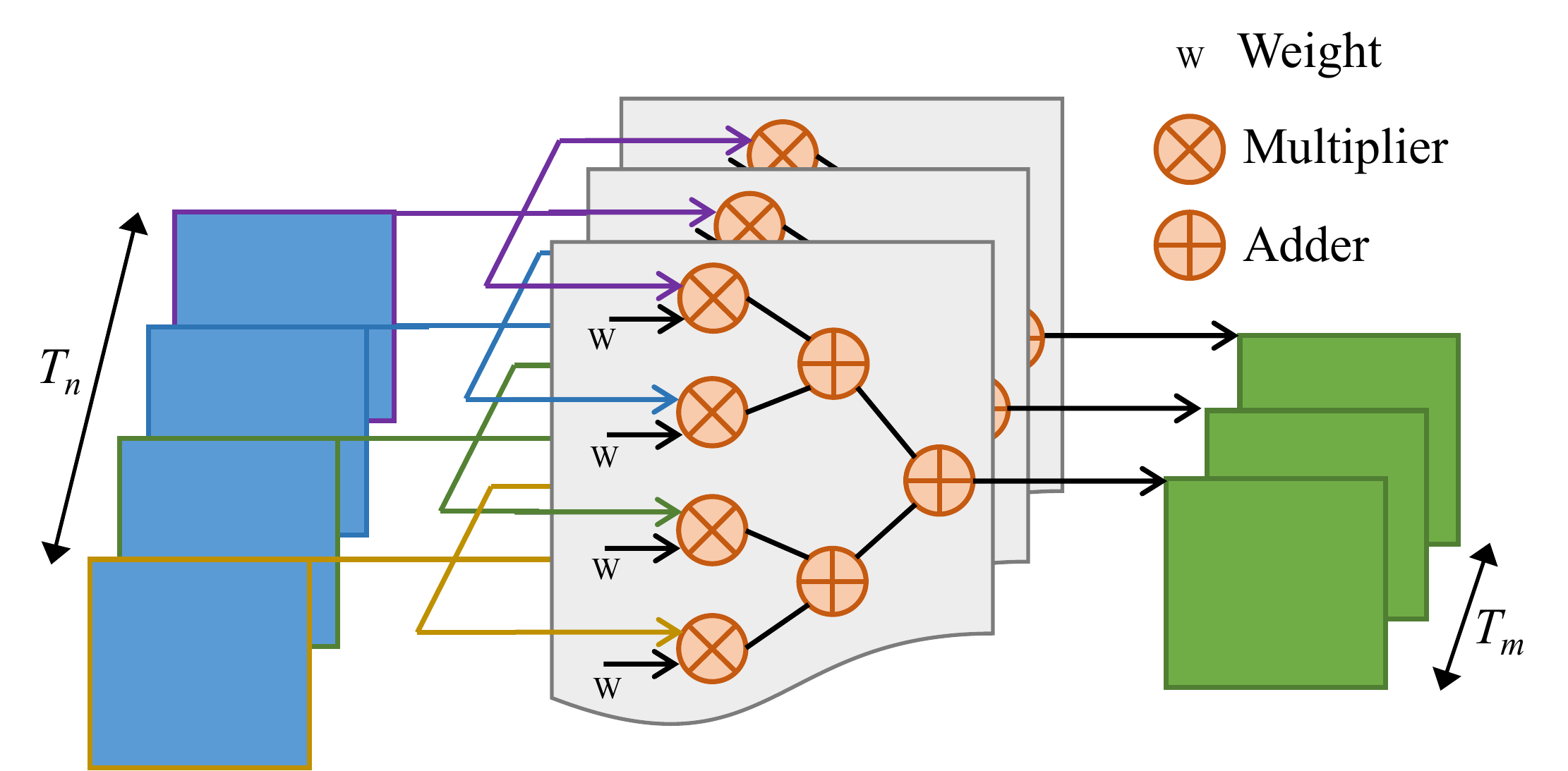}
\caption{Loop unrolling generates execution copies for parallel computation.}
\label{conv}
\end{figure} 

\textbf{Computation optimization.} Loop unrolling is pivotal for utilizing all available computational resources and maximizing computation parallelism. Taking convolution PE as an example, as illustrated in Fig.\ref{conv}, the loops of dimensions \textit{Tn} and \textit{Tm} are unrolled, generating computation instances consisting of \textit{Tm} groups with \textit{Tn} multiplications and adder trees. Each input feature map needs to be broadcasted to all instances in the \textit{Tm} dimension, allowing simultaneous computation for \textit{Tn} input feature maps.

\textbf{Communication optimization.} Within the available storage resources, it is paramount to maximize the on-chip buffer to enhance data reusability, thereby reducing the frequency and duration of data transfers.

\subsection{AGC-specific PEs}

\begin{figure}[!t]
\centering
\includegraphics[width=0.5\textwidth]{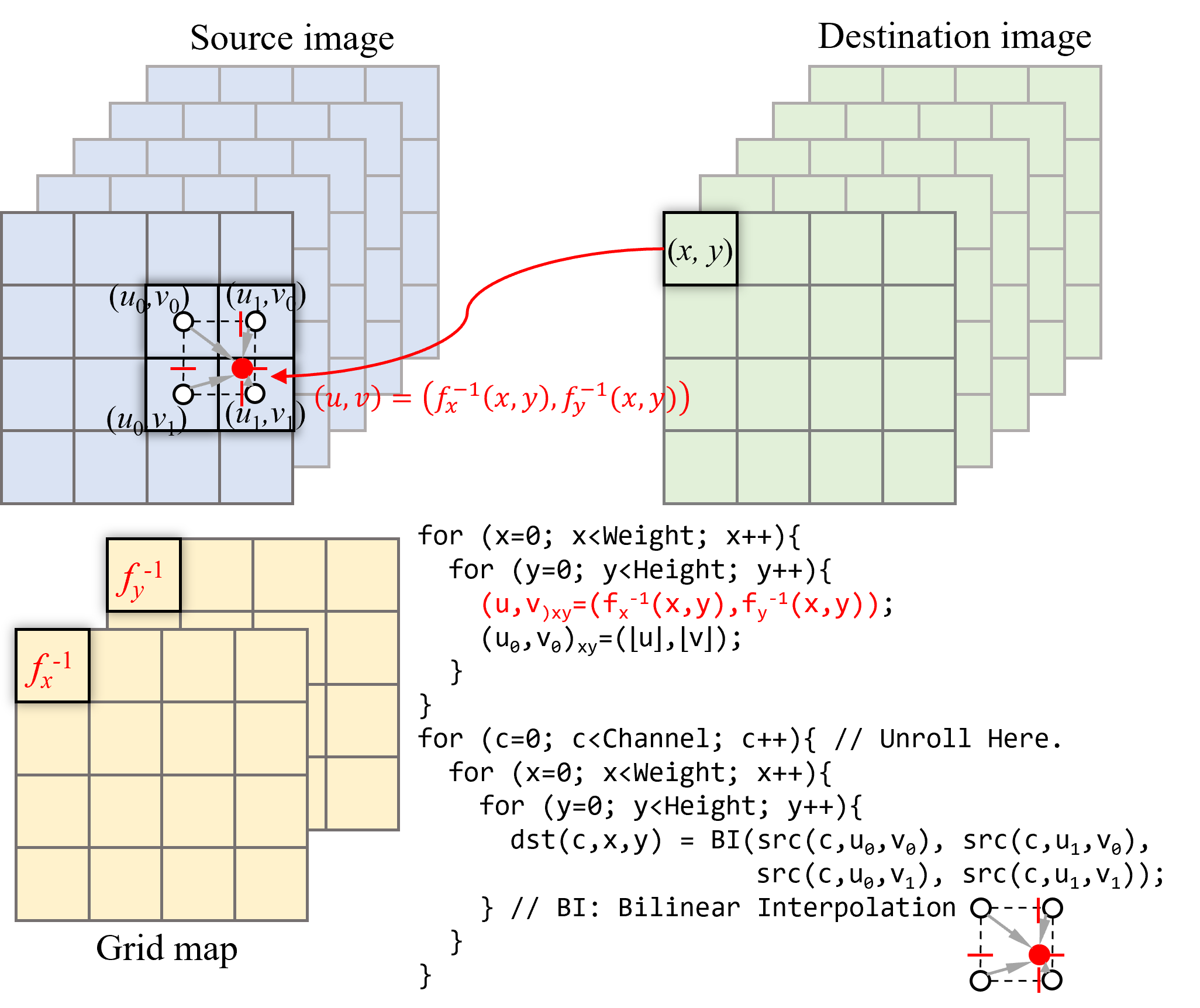}
\caption{Grid sample works as an AGC-special PE which occurs rarely in CNN accelerators.}
\label{grid_sample}
\end{figure}

The inclusion of operators such as upsampling, Hadamard Product and grid sampling is indispensable in AGC, which is not commonly observed in conventional CNN algorithms. The implementations of upsampling and Hadamard Product are relatively straightforward, requiring only adjustments to the tile size to achieve a balance between speed and resource utilization. In contrast, grid sampling stands out as the most representative specialized operator. Subsequently, its implementation will be described using it as an example.Its implementation process is illustrated in Fig.\ref{grid_sample}.

 The objective of lattice sampling is to distort the source image into the target image based on a grid map describing affine transformations. This grid has the same height and width as the destination image and contains two channels representing coordinate information in two dimensions. The semantic information in the image generally remains unchanged, allowing human eyes to still recognize it, thereby achieving the function of the "Warp" module in the AGC algorithm. Firstly, each position of the grid map is traversed, and based on the biaxial information recorded, a non-integer coordinate position (2 fraction bits within 8 bits) on the source image is mapped back. After rounding down, the nearest integer coordinate position (no fraction bits within 6 bits) to the upper-left position of the point is obtained, while the difference between the non-integer coordinates and the integer coordinates in the biaxial direction is calculated through subtraction (2 fraction bits):
\begin{equation}
    du = u - u_0, dv = v - v_0
\end{equation}
Secondly, each position of the destination image is traversed, and the pixel value at that location is calculated using bilinear interpolation, which considers the values of the four neighboring pixels on the source image:
\begin{equation}
    \label{bilinear}
    \begin{split}
        dst(x, y) = &(1-du)\times(1-dv)\times src(u_0, v_0)\\
        + &(1-du)\times dv\times src(u_0, v_0+1)\\
    + &du\times(1-dv)\times src(u_0+1, v_0) \\
    +& du\times dv\times src(u_0+1, v_0+1)
    \end{split}
\end{equation}
Due to the relatively small dimensions of the images to be processed (64×64) and a large number of channels (256), the tile size is set to 1×64×64, representing a single complete channel in the image. This choice simplifies the design while ensuring that on-chip storage space is not exceeded. Loop unrolling is performed along the tile dimension, where multiple tiles utilize the same buffer containing mapped source image coordinate information, facilitating simultaneous sampling calculations and accelerating the overall process.

\section{Experiments} \label{Exp}

We conducted comprehensive experiments to validate the effectiveness of the hardware-oriented AGC decoding and its prototype system based on FPGA. Section \ref{setup} showcases the dataset and experimental settings. Section \ref{sub} presents the comparison of compression performance among standard codecs including HEVC and VVC\footnote{We use the HEVC reference software FFmpeg x265-medium and VVC reference software VTM-10.0 in our evaluation.}, and AGC models including the float AGC (refer to~\ref{sec:baseline}) and the integer AGC (refer to~\ref{sec:integerAGC}). Section \ref{prototype_sec} demonstrates the implementation details and system-level features of the FPGA prototype.

\subsection{Experiment Setup}\label{setup}
\subsubsection{Datasets} Following \cite{konuko2021ultra}, two datasets are employed for training and testing. The \textit{VoxCeleb} dataset\cite{Nagrani_2017} contains face videos extracted from YouTube. Following the pre-processing of \cite{siarohin2019first}, original videos are filtered, cropped, and resized into 12331 training videos and 444 test videos with a resolution of 256$\times$256 and with lengths less than 1024 frames. The \textit{Xiph.org} dataset\footnote{https://media.xiph.org/video/derf/} includes 16 sequences of talking faces. Similar pre-processing is utilized on them and the results videos are used for test only.

\subsubsection{Quality Metrics} To evaluate the objective reconstruction performance, PSNR (Peak Signal-to-Noise Ratio), MS\mbox{-}SSIM (Multi-Scale Structural Similarity)~\cite{wang2003multiscale} and LPIPS (Learned Perceptual Image Patch Similarity)~\cite{zhang2018unreasonable} is adopted. PSNR is a traditional and widely-used metric focusing on pixel-wise fidelity. It quantifies the difference between the original and compressed versions by comparing their peak signal power to the power of the noise introduced during image processing. MS\mbox{-}SSIM\cite{wang2003multiscale} assesses the perceptual similarity between two images. It evaluates the luminance, contrast, and structural information at multiple scales, taking into account the human visual system's (HVS) characteristics. LPIPS\cite{zhang2018unreasonable} compares feature representations extracted from different layers of a pre-trained neural network. It measures the similarity of deep semantic features to better align with human perceptual judgments, which is often preferred in video talking scenarios where perceptual quality is a crucial consideration.

\subsubsection{Hardware Platform Configuration} To valid the effectiveness of the proposed deployment flow and system, we targeted PYNQ-Z1\footnote{https://digilent.com/reference/programmable-logic/pynq-z1/reference-manual}, a Zynq deployment board equipped with a ZYNQ XC7Z020-1CLG400C chip, comprising 650MHz ARM\textregistered Cortex\textregistered-A9 dual-core processor and programmable logics. 512MB DDR3 with 16-bit bus @ 1050Mbps is used as off-chip storage, equipped with a MicroSD slot and UART interface. The development environment is Vitis HLS 2020.1 for PL, Vivado 2020.1 for hardware platform design, and Vitis Unified IDE 2020.1 for software platform. For comparison, commercial CPU and GPU are chosen as target devices for deployment. Here we used Intel\textregistered Xeon CPU E5-2620 v4 and Nvidia Titan RTX.

\subsection{Comparison of Compression Performance}\label{sub}


\begin{figure*}[!t]
\centering
\footnotesize
\def\xheight{1}
\setlength{\tabcolsep}{1pt}
    \includegraphics[width=\xheight\linewidth]{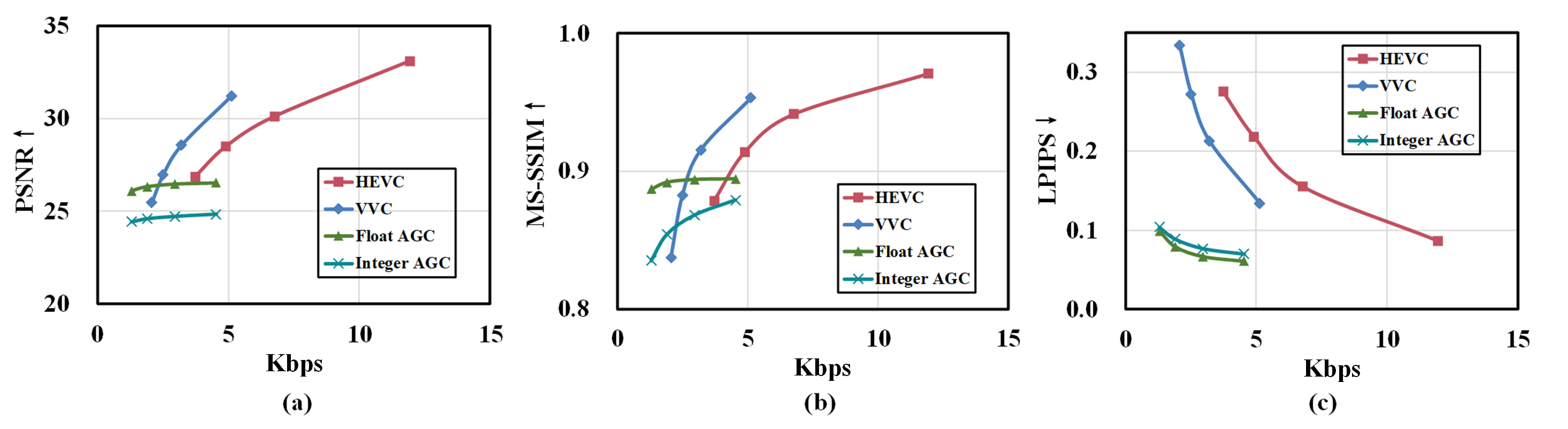}
\vspace{-0.3cm}
\caption{R-D curves of traditional codecs (HEVC and VVC) and the AGC models (float AGC and integer AGC), utilizing (a) PSNR, (b) MS\mbox{-}SSIM, and (c) LPIPS as quality evaluation metrics. The bitrate of compressed videos is calculated as kilobits per second (kbps).  PSNR and MS\mbox{-}SSIM are calculated in YUV color space while LPIPS is calculated in RGB color space.}
\label{fig:rd_curve}
\end{figure*}

\begin{table*}[]
\centering
\caption{Quality decline caused by quantification. The bitrates, PSNR, MS\mbox{-}SSIM, and LPIPS were compared for the model before quantization and after post-training static quantization under two different QPs.}
\label{quantization}
\footnotesize
{%
\begin{tabular}{c|c|c|c|c|c|c|c|c}
\hline
\textbf{}                     & \multicolumn{4}{c|}{Source QP=30}       & \multicolumn{4}{c}{Source QP=40}      \\ \cline{2-9} 
\textbf{}                     & kbps & PSNR (dB)$\uparrow$ & MS-SSIM$\uparrow$   & LPIPS$\downarrow$ & kbps & PSNR (dB)$\uparrow$ & MS-SSIM$\uparrow$ & LPIPS$\downarrow$ \\ \hline
\textbf{Float-point Baseline} &  2.958   &  26.477    &    0.894       &   0.066    &  1.306   &   26.099   &    0.886     &   0.099    \\ \hline
\textbf{After Calibration}   &  2.958   &   24.717   &    0.868     &   0.077      &   1.306  &   24.533   &    0.835     &   0.104    \\\hline
\end{tabular}%
}
\end{table*}

\begin{figure*}[!t]
\centering
\footnotesize
\def\xheight{0.85}
\setlength{\tabcolsep}{1pt}
    \includegraphics[width=\xheight\linewidth]{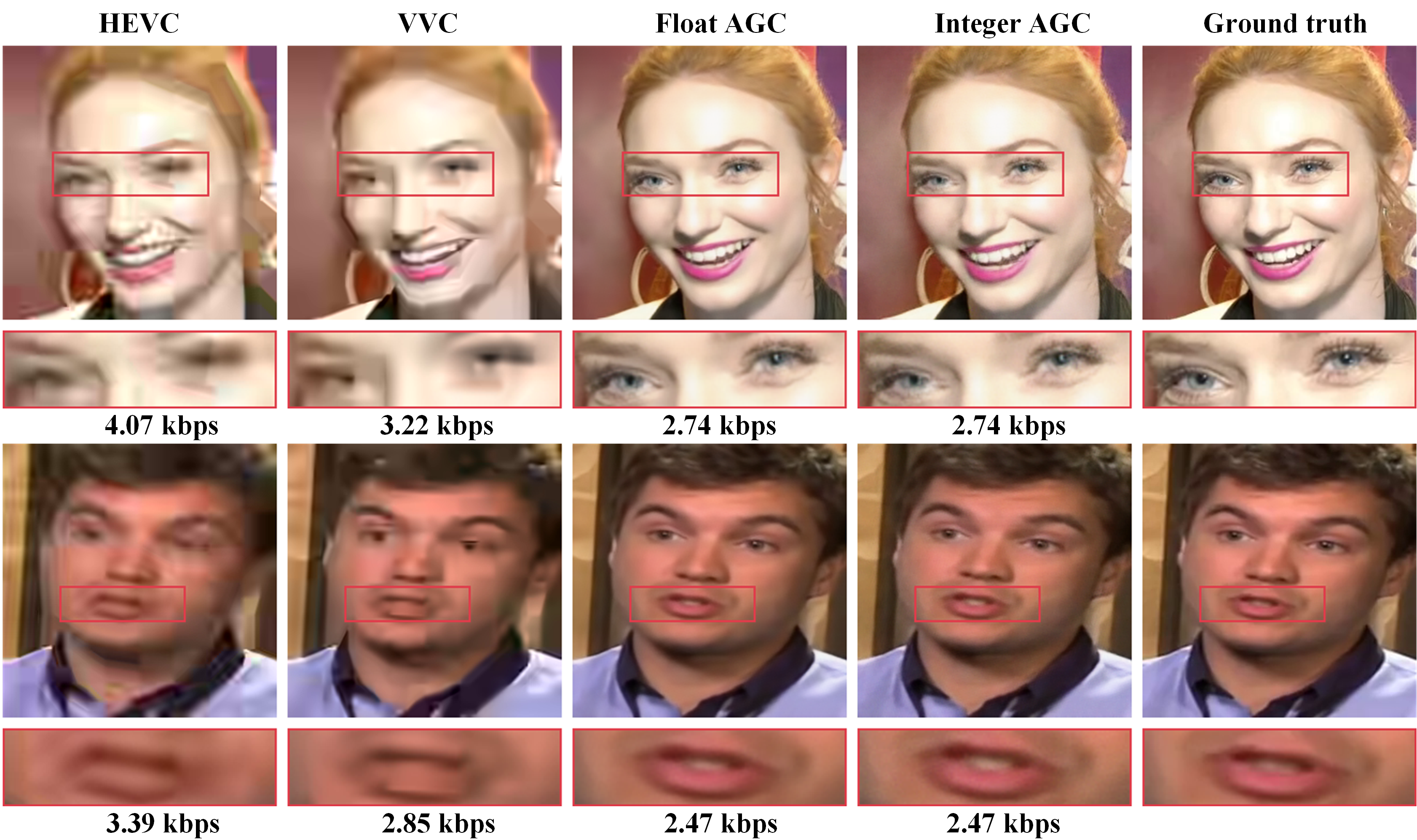}
\vspace{-0.3cm}
\caption{Qualitative results of evaluated compression frameworks including HEVC, VVC, animation-based generative codec before post-training static quantization (AGC before PTSQ), and animation-based generative codec before post-training static quantization (AGC after PTSQ), from left to right respectively. Bitrates measured by kilobits per second (kbps) of each compressed video are shown at the bottom.}
\label{fig:visual_results}
\end{figure*}


The Rate-Distortion (R-D) curves depicted in Fig.\ref{fig:rd_curve} compare high-performance traditional codecs (HEVC and VVC) with the float AGC and integer AGC. Since AGC incorporates the merit of image animation and tends to compress videos at lower bitrates, we assigned larger QPs for HEVC and VVC to facilitate comparable low-bitrate comparisons among these three compression frameworks. Specifically, we set the QPs of HEVC and VVC as (40,45,48,51) and set the QPs of source frames in our AGC models as (25,30,35,40). 

Fig.\ref{fig:rd_curve} (a), (b), and (c) utilize PSNR, MS\mbox{-}SSIM, and LPIPS as quality evaluation metrics, respectively. The arrows in the legend of the Y-axis represent the trend of numerical transformations of quality metrics corresponding to better reconstruction quality of compressed videos. For example, a downward arrow for LPIPS indicates that lower LPIPS values represent better reconstruction quality. Note that while PSNR calculates the pixel-level distortion and MS-SSIM compares the structural similarity, the calculation of both is based on pixel values. As for LPIPS, on the other hand, its use of deep learning models trained on large-scale image datasets to mimic human perception allows it to capture complex perceptual differences that traditional metrics may overlook. As a result, LPIPS often provides more reliable and consistent assessments of image similarity and quality. The curves illustrate that, in comparison to conventional codecs, the AGC models exhibit superior reconstruction MS\mbox{-}SSIM and LPIPS at ultra-low bitrates (lower than 5 kbps). However, AGC exhibits a slightly inferior performance in terms of PSNR. In other words, the perceptual quality superiority of AGC comes at the cost of a reduction in fidelity, aligning with the well-known rate-distortion-perception theory\cite{blau2018perception,blau2019rethinking,zhang2021universal}, which is a common limitation in generative models. It is worth noting that the integer AGC has a negligible performance gap against the float counterpart, especially for LPIPS focusing on perceptual quality.

Table \ref{quantization} presents the degradation in reconstruction quality introduced by quantization at QP=30 and QP=40 for source frames. The data in the table indicates that quantization unavoidably introduces a decline in all three metrics, albeit by a mere -1.66dB, -0.077, and +0.008 for PSNR, MS\mbox{-}SSIM, and LPIPS, respectively for average at two QPs. This level of quality loss is deemed acceptable for edge devices in low-power scenarios. Furthermore, in scenarios where the AGC performs optimally at ultra-low bit rates, users exhibit tolerance towards minor losses in video quality. Since the quantization process is exclusive to the decoder, it is imperative to note that the bitrate, as determined by the encoding end, remains constant and undergoes no variation.

In addition to the RD curves, sampled frames of the reconstruction results for three test videos are presented for both traditional codecs and AGC models. As illustrated in Fig.~\ref{fig:visual_results}, at ultra-low bitrates (below 5 kbps), the videos reconstructed by AGC models exhibit superior visual quality compared to the reconstruction outputs of HEVC and VVC. The contours of facial features are more distinct and smooth in AGC-reconstructed videos, while traditional codec-reconstructed videos at this bitrate display evident artifacts such as blurring and block effects. Furthermore, the videos reconstructed by the integer AGC show a minimal perceptual difference compared to the counterparts reconstructed by the float AGC, with subtle granularity discernible upon close examination.

\subsection{FPGA Prototype} \label{prototype_sec}

\begin{figure}[!h]
\centering
\includegraphics[width=0.4\textwidth]{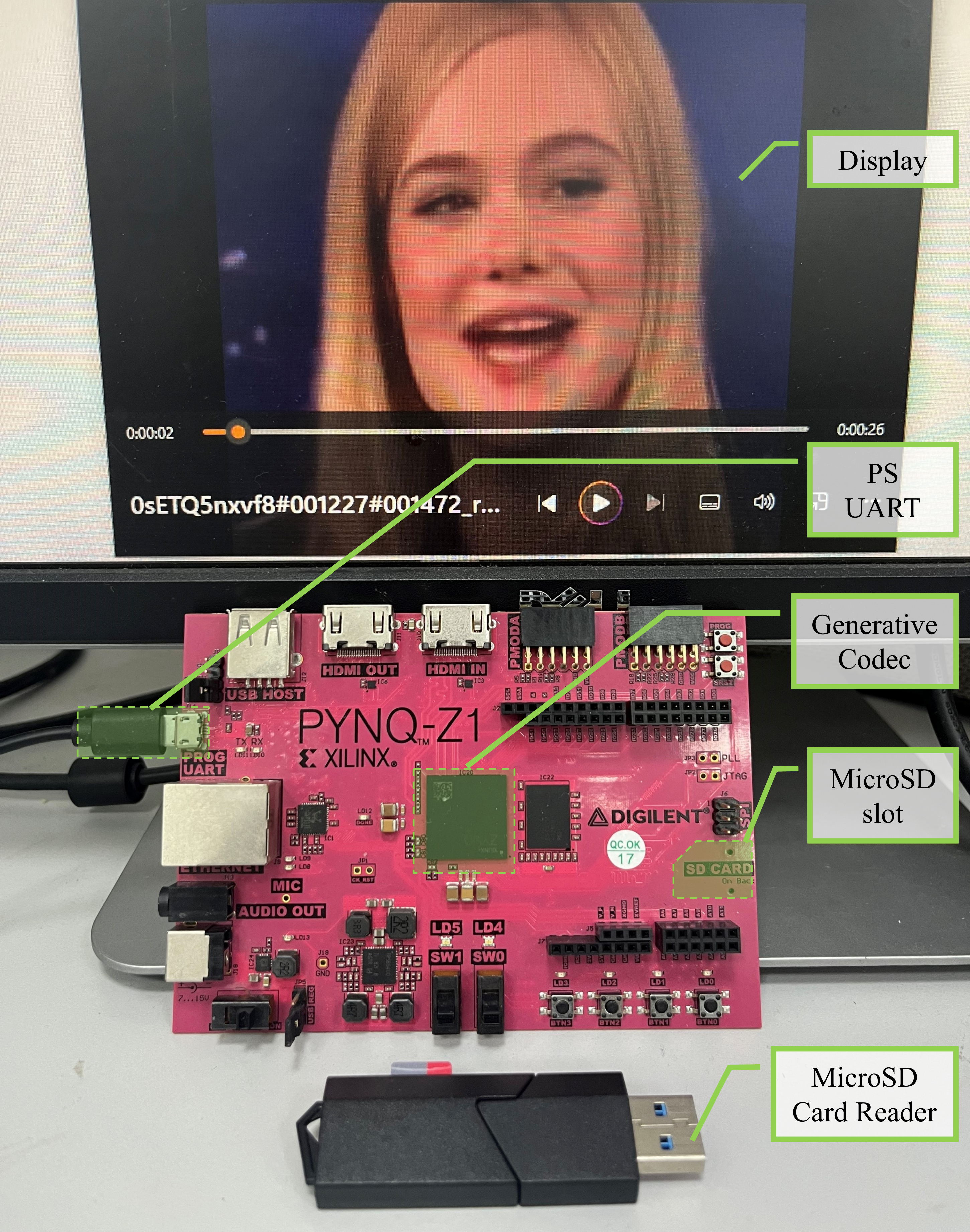}
\caption{The picture of the proposed prototype. In addition to the FPGA board, a MicroSD card and a card reader have been employed as the data transmission pathway between the FPGA system and the display device.}
\label{prototype}
\end{figure}

\begin{figure}[!h]
\centering
\includegraphics[width=0.5\textwidth]{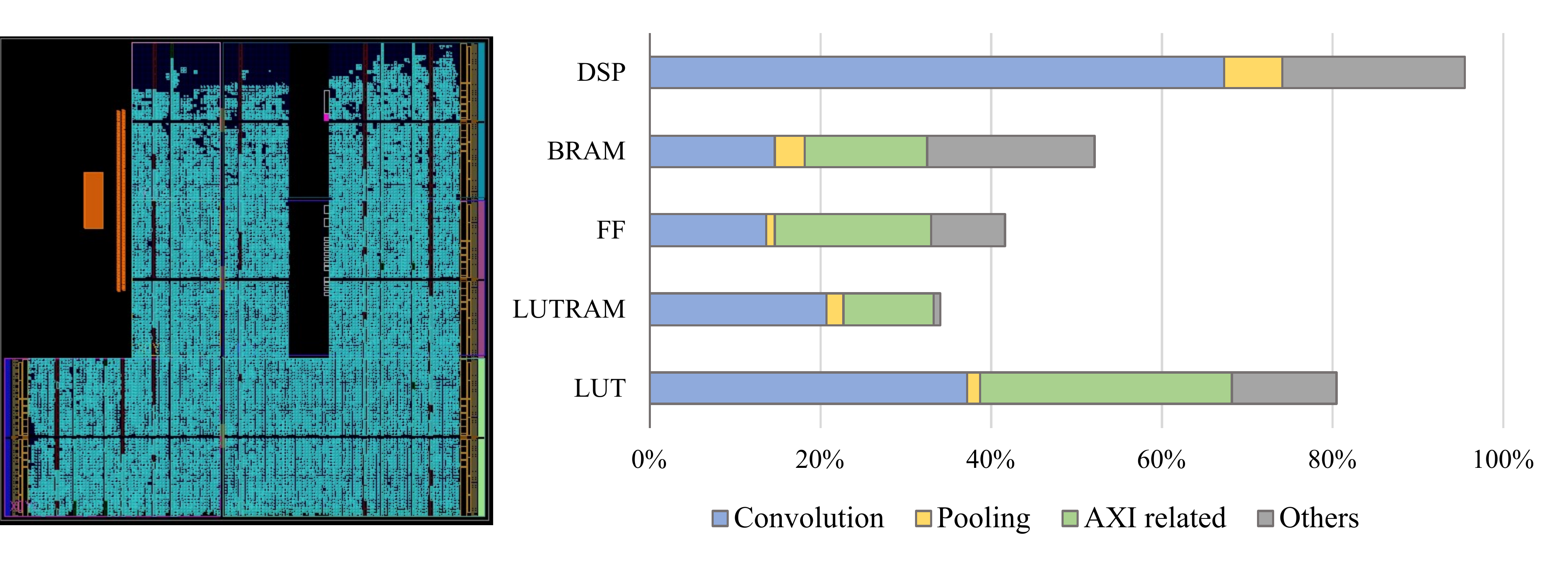}
\caption{The place-and-routed result and resource utilization.}
\label{place}
\end{figure}

\begin{table*}[htbp]
\centering
\caption{Module-wise FPGA Resource Utilization}
\label{tab:module_resource_util}
\begin{tabular}{c|c|c|c|c|c}
\hline
\textbf{Module} & \textbf{LUT} & \textbf{LUTRAM} & \textbf{FF} & \textbf{BRAM} & \textbf{DSP} \\
\hline
Convolution & 12842 (30.00\%) & 2100 (35.46\%) & 18000 (40.65\%) & 90 (70.00\%) & 148 (70.48\%) \\
Pooling     & 2140 (5.00\%)   & 350 (5.91\%)   & 2100 (4.74\%)   & 6  (4.67\%)  & 12  (5.71\%)  \\
AXI related & 12842 (30.00\%) & 1480 (25.00\%) & 13285 (30.00\%) & 16 (12.40\%) & 0 (0.00\%)   \\
Others      & 15000 (35.00\%) & 1992 (33.63\%) & 11000 (24.85\%) & 16 (12.40\%) & 50 (23.81\%) \\
\hline
Overall     & 42806 (80.46\%) & 5922 (34.03\%) & 44284 (41.62\%) & 129 (52.14\%) & 210 (95.45\%) \\
\hline
\end{tabular}
\end{table*}

\begin{table}[t]
\centering
\caption{Resource utilization of the FPGA system}
\label{utilization}
\resizebox{\columnwidth}{!}{%
\begin{tabular}{c|ccccc}
\hline
Resource          & DSP  & BRAM(KB) & FF  & LUTRAM  & LUT   \\ \hline
Utilization       & 210   & 328.48   & 44284  & 5922      & 42806   \\
Utilization (\%)  & 95.45   & 52.14   & 41.62  & 34.03    & 80.46      \\
\hline
\end{tabular}%
}
\vspace{-0.3cm}
\end{table}


\begin{table}[t]
\centering
\caption{Performance comparison between CPU, GPU, and FPGA platforms, with the video resolution of 256$\times$256}
\label{platform_comp}
\begin{threeparttable}
\resizebox{\columnwidth}{!}{%
\begin{tabular}{c|cc|c}
\hline
\multicolumn{1}{c|}{Platform} & \multicolumn{2}{c|}{Jetson Xavier NX} & PYNQ-Z1 \\ \cline{2-4}
                              & CPU & GPU & XC7Z020 \\ \hline
Precision                     & FP32 & FP32 & Mix(INT8+FP32) \\
Throughput (FPS)\,($\uparrow$)            & 0.12 & 2.21 & \textbf{2.38} \\
Power (W)\,($\downarrow$)                 & 6.7  & 10.1 & \textbf{1.83} \\
Energy Efficiency (Frame/J)\,($\uparrow$) & 0.052 & 0.321 & \textbf{1.30} \\ \hline
\end{tabular}%
}
\begin{tablenotes}\footnotesize
\item[1] Throughput/power (Frame/J).
\end{tablenotes}
\end{threeparttable}
\vspace{-0.3cm}
\end{table}

Fig.\ref{prototype} shows a picture of the proposed prototype, which could be regarded as a special instance of deep generative codec accelerated by FPGA-based heterogeneous computing. The prototype consists of the FPGA system, video display, MicroSD card with a card reader, and a host for launching the driver program in PS (not depicted in the illustration). The PS-PL-based system for the deep generative codec is deployed in the FPGA platform. The PS UART control can be connected to the appropriate MIO pins to control the MicroSD port. MicroSD has been employed as the data transmission pathway between the FPGA system and the display device, and consideration may be given in the future to the substitution of more efficient methodologies. Furthermore, the transmission 
 of bitstreams and weights for the network to the FPGA is also facilitated through the utilization of MicroSD.

The resource utilization in the FPGA device are shown in Table \ref{utilization}. The prototype system has consumed 42 806 Slice LUTs and 44 284 Slice Registers(\textit{e.g.} Flip-Flops (FFs) in FPGA) that correspond to 80\% of total Slice LUTs and 42\% of total Slice Registers. Furthermore, 328 KB on-chip memory(\textit{e.g.} BRAM in FPGA) has been used. Besides, the system has almost fully used 220 DSP blocks in the FPGA device. From Fig.\ref{place}, it can be observed that convolutional operations consume a significant portion of DSP resources, along with a portion of logic and storage resources. This is attributed to the fact that convolutional operations constitute a fundamental component of AGC tasks, and the large size and frequent occurrences of these operations ensure the potent generative capabilities of AGC, making them the most time-consuming tasks requiring a substantial allocation of computational resources. Additionally, the AXI bus, serving as the protocol for data exchange between the PS and PL, also occupies a considerable amount of resources. Furthermore, there is a substantial amount of unused Slices and BRAMs on the FPGA, providing optimization potential for further performance enhancement in our system.

The system-level features of the system are shown in Table \ref{platform_comp}. The system, running in XC7Z020 at 100 MHz, decodes a 256$\times$256 video with a throughput of 2.38 frames per section (FPS). To validate the efficiency of the proposed system, the AGC decoder without model compression is implemented on Intel\textregistered Xeon\textregistered CPU and Nvidia Titan RTX GPU. It should be noted that the table exclusively enumerates the task characteristics on the PL side. This is due to the fact that tasks on the PS side are uniformly executed on the CPU across all platforms and, as such, have been excluded from the scope of the statistics. The table reveals that our system achieves throughput close to that of the CPU at extremely low power consumption, with an energy efficiency (Frame\/s\/W or Frame\/J) ratio of 24.9$\times$ that of the CPU and 4.06$\times$ that of the GPU. Specifically, our system only requires 11.7 $\upmu$J to produce a one-pixel result, compared to 47.6 $\upmu$J for the GPU and 292 $\upmu$J for the CPU.

To the best of our knowledge, we are the first to implement animation-based generative talking-face video decoding on the FPGA platform, showing potential in the deployment of future video compression algorithms on embedded systems or edge devices.

\section{Future work} \label{future}
Despite experimental evidence demonstrating the potential performance of our pioneering AGC system, additional and more in-depth research is warranted. Primarily, the PYNQ-Z1 used here has limited resources, while resource-rich development boards are imperative for applications in deep learning. Furthermore, as an inaugural project developing the FPGA system tailored for AGC applications, many AGC-specific operators have been analyzed exhaustively and implemented functionally correctly with proper resource utilization. In future endeavors, we anticipate the deployment of this project on resource-rich development boards, showcasing enhanced capabilities through optimized parallelism. Beyond experimental development boards, we also envision the deployment of this system on authentic edge devices, thereby demonstrating its significant application potential in industrial settings.

\section{Conclusion} \label{Conclusion}
This paper presents a hardware-software co-design processing system for the AGC decoder. We initially analyze AGC methods, with a specific focus on the decoder possessing robust generative capabilities. To deploy this decoder on resource-constrained hardware platforms, we employ quantization and layer fusion techniques for network compression, resulting in a 0.25$\times$ reduction in model size. Our integer AGC achieves superior visual quality in the reconstruction of videos at less than 5 kbps compared to traditional codecs. Leveraging the proposed generic deployment flow, we establish a hardware-based AGC decoding system within an SoC-based FPGA. The prototype system demonstrates that FPGAs are promising candidates for efficient AGC decoders in edge devices, achieving 24.9$\times$ and 4.1$\times$ higher energy efficiency against commercial CPU and GPU.




\begin{thebibliography}{10}
\providecommand{\url}[1]{#1}
\csname url@samestyle\endcsname
\providecommand{\newblock}{\relax}
\providecommand{\bibinfo}[2]{#2}
\providecommand{\BIBentrySTDinterwordspacing}{\spaceskip=0pt\relax}
\providecommand{\BIBentryALTinterwordstretchfactor}{4}
\providecommand{\BIBentryALTinterwordspacing}{\spaceskip=\fontdimen2\font plus
\BIBentryALTinterwordstretchfactor\fontdimen3\font minus \fontdimen4\font\relax}
\providecommand{\BIBforeignlanguage}[2]{{%
\expandafter\ifx\csname l@#1\endcsname\relax
\typeout{** WARNING: IEEEtran.bst: No hyphenation pattern has been}%
\typeout{** loaded for the language `#1'. Using the pattern for}%
\typeout{** the default language instead.}%
\else
\language=\csname l@#1\endcsname
\fi
#2}}
\providecommand{\BIBdecl}{\relax}
\BIBdecl

\bibitem{balle2016end}
J.~Ball{\'e}, V.~Laparra, and E.~P. Simoncelli, ``End-to-end optimized image compression,'' \emph{arXiv preprint arXiv:1611.01704}, 2016.

\bibitem{balle2018variational}
J.~Ball{\'e}, D.~Minnen, S.~Singh, S.~J. Hwang, and N.~Johnston, ``Variational image compression with a scale hyperprior,'' \emph{arXiv preprint arXiv:1802.01436}, 2018.

\bibitem{chen2022exploiting}
Z.~Chen, S.~Gu, G.~Lu, and D.~Xu, ``Exploiting intra-slice and inter-slice redundancy for learning-based lossless volumetric image compression,'' \emph{IEEE Transactions on Image Processing}, vol.~31, pp. 1697--1707, 2022.

\bibitem{theis2017lossy}
L.~Theis, W.~Shi, A.~Cunningham, and F.~Husz{\'a}r, ``Lossy image compression with compressive autoencoders,'' \emph{arXiv preprint arXiv:1703.00395}, 2017.

\bibitem{wu2018video}
C.-Y. Wu, N.~Singhal, and P.~Krahenbuhl, ``Video compression through image interpolation,'' in \emph{Proceedings of the European conference on computer vision (ECCV)}, 2018, pp. 416--431.

\bibitem{minnen2018joint}
D.~Minnen, J.~Ball{\'e}, and G.~D. Toderici, ``Joint autoregressive and hierarchical priors for learned image compression,'' \emph{Advances in neural information processing systems}, vol.~31, 2018.

\bibitem{wang2021one}
T.-C. Wang, A.~Mallya, and M.-Y. Liu, ``One-shot free-view neural talking-head synthesis for video conferencing,'' in \emph{Proceedings of the IEEE/CVF conference on computer vision and pattern recognition}, 2021, pp. 10\,039--10\,049.

\bibitem{oquab2021low}
M.~Oquab, P.~Stock, D.~Haziza, T.~Xu, P.~Zhang, O.~Celebi, Y.~Hasson, P.~Labatut, B.~Bose-Kolanu, T.~Peyronel \emph{et~al.}, ``Low bandwidth video-chat compression using deep generative models,'' in \emph{Proceedings of the IEEE/CVF Conference on Computer Vision and Pattern Recognition}, 2021, pp. 2388--2397.

\bibitem{konuko2021ultra}
G.~Konuko, G.~Valenzise, and S.~Lathuili{\`e}re, ``Ultra-low bitrate video conferencing using deep image animation,'' in \emph{ICASSP 2021-2021 IEEE International Conference on Acoustics, Speech and Signal Processing (ICASSP)}.\hskip 1em plus 0.5em minus 0.4em\relax IEEE, 2021, pp. 4210--4214.

\bibitem{feng2021generative}
D.~Feng, Y.~Huang, Y.~Zhang, J.~Ling, A.~Tang, and L.~Song, ``A generative compression framework for low bandwidth video conference,'' in \emph{2021 IEEE International Conference on Multimedia \& Expo Workshops (ICMEW)}.\hskip 1em plus 0.5em minus 0.4em\relax IEEE, 2021, pp. 1--6.

\bibitem{lu2019dvc}
G.~Lu, W.~Ouyang, D.~Xu, X.~Zhang, C.~Cai, and Z.~Gao, ``Dvc: An end-to-end deep video compression framework,'' in \emph{Proceedings of the IEEE/CVF Conference on Computer Vision and Pattern Recognition}, 2019, pp. 11\,006--11\,015.

\bibitem{x265}
G.~J. Sullivan, J.-R. Ohm, W.-J. Han, and T.~Wiegand, ``Overview of the high efficiency video coding (hevc) standard,'' \emph{IEEE Transactions on circuits and systems for video technology}, vol.~22, no.~12, pp. 1649--1668, 2012.

\bibitem{SIP-2021-0044}
\BIBentryALTinterwordspacing
N.~Ling, C.-C.~J. Kuo, G.~J. Sullivan, D.~Xu, S.~Liu, H.-M. Hang, W.-H. Peng, and J.~Liu, ``The future of video coding,'' \emph{APSIPA Transactions on Signal and Information Processing}, vol.~11, no.~1, pp.~--, 2022. [Online]. Available: \url{http://dx.doi.org/10.1561/116.00000044}
\BIBentrySTDinterwordspacing

\bibitem{10044587}
J.~Song, X.~Tang, X.~Qiao, Y.~Wang, R.~Wang, and R.~Huang, ``A 28 nm 16 kb bit-scalable charge-domain transpose 6t sram in-memory computing macro,'' \emph{IEEE Transactions on Circuits and Systems I: Regular Papers}, vol.~70, no.~5, pp. 1835--1845, 2023.

\bibitem{han2015deep}
S.~Han, H.~Mao, and W.~J. Dally, ``Deep compression: Compressing deep neural networks with pruning, trained quantization and huffman coding,'' \emph{arXiv preprint arXiv:1510.00149}, 2015.

\bibitem{han2015learning}
S.~Han, J.~Pool, J.~Tran, and W.~Dally, ``Learning both weights and connections for efficient neural network,'' \emph{Advances in neural information processing systems}, vol.~28, 2015.

\bibitem{han2017ese}
S.~Han, J.~Kang, H.~Mao, Y.~Hu, X.~Li, Y.~Li, D.~Xie, H.~Luo, S.~Yao, Y.~Wang \emph{et~al.}, ``Ese: Efficient speech recognition engine with sparse lstm on fpga,'' in \emph{Proceedings of the 2017 ACM/SIGDA International Symposium on Field-Programmable Gate Arrays}, 2017, pp. 75--84.

\bibitem{vasudevan2017parallel}
A.~Vasudevan, A.~Anderson, and D.~Gregg, ``Parallel multi channel convolution using general matrix multiplication,'' in \emph{2017 IEEE 28th international conference on application-specific systems, architectures and processors (ASAP)}.\hskip 1em plus 0.5em minus 0.4em\relax IEEE, 2017, pp. 19--24.

\bibitem{chen2014dadiannao}
Y.~Chen, T.~Luo, S.~Liu, S.~Zhang, L.~He, J.~Wang, L.~Li, T.~Chen, Z.~Xu, N.~Sun \emph{et~al.}, ``Dadiannao: A machine-learning supercomputer,'' in \emph{2014 47th Annual IEEE/ACM International Symposium on Microarchitecture}.\hskip 1em plus 0.5em minus 0.4em\relax IEEE, 2014, pp. 609--622.

\bibitem{conv2016}
\BIBentryALTinterwordspacing
S.~K. Esser, P.~A. Merolla, J.~V. Arthur, A.~S. Cassidy, R.~Appuswamy, A.~Andreopoulos, D.~J. Berg, J.~L. McKinstry, T.~Melano, D.~R. Barch, C.~di~Nolfo, P.~Datta, A.~Amir, B.~Taba, M.~D. Flickner, and D.~S. Modha, ``Convolutional networks for fast, energy-efficient neuromorphic computing,'' \emph{Proceedings of the National Academy of Sciences}, vol. 113, no.~41, pp. 11\,441--11\,446, 2016. [Online]. Available: \url{https://www.pnas.org/doi/abs/10.1073/pnas.1604850113}
\BIBentrySTDinterwordspacing

\bibitem{han2016eie}
S.~Han, X.~Liu, H.~Mao, J.~Pu, A.~Pedram, M.~A. Horowitz, and W.~J. Dally, ``Eie: Efficient inference engine on compressed deep neural network,'' \emph{ACM SIGARCH Computer Architecture News}, vol.~44, no.~3, pp. 243--254, 2016.

\bibitem{qiu2016going}
J.~Qiu, J.~Wang, S.~Yao, K.~Guo, B.~Li, E.~Zhou, J.~Yu, T.~Tang, N.~Xu, S.~Song \emph{et~al.}, ``Going deeper with embedded fpga platform for convolutional neural network,'' in \emph{Proceedings of the 2016 ACM/SIGDA international symposium on field-programmable gate arrays}, 2016, pp. 26--35.

\bibitem{zhang2015optimizing}
C.~Zhang, P.~Li, G.~Sun, Y.~Guan, B.~Xiao, and J.~Cong, ``Optimizing fpga-based accelerator design for deep convolutional neural networks,'' in \emph{Proceedings of the 2015 ACM/SIGDA international symposium on field-programmable gate arrays}, 2015, pp. 161--170.

\bibitem{putnam2014reconfigurable}
A.~Putnam, A.~M. Caulfield, E.~S. Chung, D.~Chiou, K.~Constantinides, J.~Demme, H.~Esmaeilzadeh, J.~Fowers, G.~P. Gopal, J.~Gray \emph{et~al.}, ``A reconfigurable fabric for accelerating large-scale datacenter services,'' \emph{ACM SIGARCH Computer Architecture News}, vol.~42, no.~3, pp. 13--24, 2014.

\bibitem{8594633}
A.~Shawahna, S.~M. Sait, and A.~El-Maleh, ``Fpga-based accelerators of deep learning networks for learning and classification: A review,'' \emph{IEEE Access}, vol.~7, pp. 7823--7859, 2019.

\bibitem{vanderbauwhede2013high}
W.~Vanderbauwhede and K.~Benkrid, \emph{High-performance computing using FPGAs}.\hskip 1em plus 0.5em minus 0.4em\relax Springer, 2013, vol.~3.

\bibitem{nurvitadhi2017can}
E.~Nurvitadhi, G.~Venkatesh, J.~Sim, D.~Marr, R.~Huang, J.~Ong Gee~Hock, Y.~T. Liew, K.~Srivatsan, D.~Moss, S.~Subhaschandra \emph{et~al.}, ``Can fpgas beat gpus in accelerating next-generation deep neural networks?'' in \emph{Proceedings of the 2017 ACM/SIGDA international symposium on field-programmable gate arrays}, 2017, pp. 5--14.

\bibitem{canis2011legup}
A.~Canis, J.~Choi, M.~Aldham, V.~Zhang, A.~Kammoona, J.~H. Anderson, S.~Brown, and T.~Czajkowski, ``Legup: high-level synthesis for fpga-based processor/accelerator systems,'' in \emph{Proceedings of the 19th ACM/SIGDA international symposium on Field programmable gate arrays}, 2011, pp. 33--36.

\bibitem{cong2011high}
J.~Cong, B.~Liu, S.~Neuendorffer, J.~Noguera, K.~Vissers, and Z.~Zhang, ``High-level synthesis for fpgas: From prototyping to deployment,'' \emph{IEEE Transactions on Computer-Aided Design of Integrated Circuits and Systems}, vol.~30, no.~4, pp. 473--491, 2011.

\bibitem{liang2012high}
Y.~Liang, K.~Rupnow, Y.~Li, D.~Min, M.~N. Do, and D.~Chen, ``High-level synthesis: productivity, performance, and software constraints,'' \emph{Journal of Electrical and Computer Engineering}, vol. 2012, pp. 1--1, 2012.

\bibitem{chen2016eyeriss}
Y.-H. Chen, J.~Emer, and V.~Sze, ``Eyeriss: A spatial architecture for energy-efficient dataflow for convolutional neural networks,'' \emph{ACM SIGARCH computer architecture news}, vol.~44, no.~3, pp. 367--379, 2016.

\bibitem{DBLP:journals/corr/abs-1903-06630}
\BIBentryALTinterwordspacing
G.~G.~F. Lemieux, J.~Edwards, J.~Vandergriendt, A.~Severance, R.~D. Iaco, A.~Raouf, H.~Osman, T.~Watzka, and S.~Singh, ``Tinbinn: Tiny binarized neural network overlay in about 5, 000 4-luts and 5mw,'' \emph{CoRR}, vol. abs/1903.06630, 2019. [Online]. Available: \url{http://arxiv.org/abs/1903.06630}
\BIBentrySTDinterwordspacing

\bibitem{umuroglu2017finn}
Y.~Umuroglu, N.~J. Fraser, G.~Gambardella, M.~Blott, P.~Leong, M.~Jahre, and K.~Vissers, ``Finn: A framework for fast, scalable binarized neural network inference,'' in \emph{Proceedings of the 2017 ACM/SIGDA international symposium on field-programmable gate arrays}, 2017, pp. 65--74.

\bibitem{7544745}
S.~I. Venieris and C.-S. Bouganis, ``fpgaconvnet: A framework for mapping convolutional neural networks on fpgas,'' in \emph{2016 IEEE 24th Annual International Symposium on Field-Programmable Custom Computing Machines (FCCM)}, 2016, pp. 40--47.

\bibitem{7827589}
C.~Zhang, Z.~Fang, P.~Zhou, P.~Pan, and J.~Cong, ``Caffeine: Towards uniformed representation and acceleration for deep convolutional neural networks,'' in \emph{2016 IEEE/ACM International Conference on Computer-Aided Design (ICCAD)}, 2016, pp. 1--8.

\bibitem{9415618}
L.~Petrica, T.~Alonso, M.~Kroes, N.~Fraser, S.~Cotofana, and M.~Blott, ``Memory-efficient dataflow inference for deep cnns on fpga,'' in \emph{2020 International Conference on Field-Programmable Technology (ICFPT)}, 2020, pp. 48--55.

\bibitem{cadambi2010programmable}
S.~Cadambi, A.~Majumdar, M.~Becchi, S.~Chakradhar, and H.~P. Graf, ``A programmable parallel accelerator for learning and classification,'' in \emph{Proceedings of the 19th international conference on Parallel architectures and compilation techniques}, 2010, pp. 273--284.

\bibitem{ovtcharov2015accelerating}
K.~Ovtcharov, O.~Ruwase, J.-Y. Kim, J.~Fowers, K.~Strauss, and E.~S. Chung, ``Accelerating deep convolutional neural networks using specialized hardware,'' \emph{Microsoft Research Whitepaper}, vol.~2, no.~11, pp. 1--4, 2015.

\bibitem{farabet2009cnp}
C.~Farabet, C.~Poulet, J.~Y. Han, and Y.~LeCun, ``Cnp: An fpga-based processor for convolutional networks,'' in \emph{2009 International Conference on Field Programmable Logic and Applications}.\hskip 1em plus 0.5em minus 0.4em\relax IEEE, 2009, pp. 32--37.

\bibitem{sankaradas2009massively}
M.~Sankaradas, V.~Jakkula, S.~Cadambi, S.~Chakradhar, I.~Durdanovic, E.~Cosatto, and H.~P. Graf, ``A massively parallel coprocessor for convolutional neural networks,'' in \emph{2009 20th IEEE International Conference on Application-specific Systems, Architectures and Processors}.\hskip 1em plus 0.5em minus 0.4em\relax IEEE, 2009, pp. 53--60.

\bibitem{chakradhar2010dynamically}
S.~Chakradhar, M.~Sankaradas, V.~Jakkula, and S.~Cadambi, ``A dynamically configurable coprocessor for convolutional neural networks,'' in \emph{Proceedings of the 37th annual international symposium on Computer architecture}, 2010, pp. 247--257.

\bibitem{gokhale2014240}
V.~Gokhale, J.~Jin, A.~Dundar, B.~Martini, and E.~Culurciello, ``A 240 g-ops/s mobile coprocessor for deep neural networks,'' in \emph{Proceedings of the IEEE conference on computer vision and pattern recognition workshops}, 2014, pp. 682--687.

\bibitem{bross2021overview}
B.~Bross, Y.-K. Wang, Y.~Ye, S.~Liu, J.~Chen, G.~J. Sullivan, and J.-R. Ohm, ``Overview of the versatile video coding (vvc) standard and its applications,'' \emph{IEEE Transactions on Circuits and Systems for Video Technology}, vol.~31, no.~10, pp. 3736--3764, 2021.

\bibitem{sullivan2012overview}
G.~J. Sullivan, J.-R. Ohm, W.-J. Han, and T.~Wiegand, ``Overview of the high efficiency video coding (hevc) standard,'' \emph{IEEE Transactions on circuits and systems for video technology}, vol.~22, no.~12, pp. 1649--1668, 2012.

\bibitem{han2021technical}
J.~Han, B.~Li, D.~Mukherjee, C.-H. Chiang, A.~Grange, C.~Chen, H.~Su, S.~Parker, S.~Deng, U.~Joshi \emph{et~al.}, ``A technical overview of av1,'' \emph{Proceedings of the IEEE}, vol. 109, no.~9, pp. 1435--1462, 2021.

\bibitem{mukherjee2015technical}
D.~Mukherjee, J.~Han, J.~Bankoski, R.~Bultje, A.~Grange, J.~Koleszar, P.~Wilkins, and Y.~Xu, ``A technical overview of vp9—the latest open-source video codec,'' \emph{SMPTE Motion Imaging Journal}, vol. 124, no.~1, pp. 44--54, 2015.

\bibitem{zhang2019recent}
J.~Zhang, C.~Jia, M.~Lei, S.~Wang, S.~Ma, and W.~Gao, ``Recent development of avs video coding standard: Avs3,'' in \emph{2019 picture coding symposium (PCS)}.\hskip 1em plus 0.5em minus 0.4em\relax IEEE, 2019, pp. 1--5.

\bibitem{gao2014overview}
W.~Gao, S.~Ma, W.~Gao, and S.~Ma, ``An overview of avs2 standard,'' \emph{Advanced Video Coding Systems}, pp. 35--49, 2014.

\bibitem{zhao2018enhanced}
Z.~Zhao, S.~Wang, S.~Wang, X.~Zhang, S.~Ma, and J.~Yang, ``Enhanced bi-prediction with convolutional neural network for high-efficiency video coding,'' \emph{IEEE Transactions on Circuits and Systems for Video Technology}, vol.~29, no.~11, pp. 3291--3301, 2018.

\bibitem{zhang2018residual}
Y.~Zhang, T.~Shen, X.~Ji, Y.~Zhang, R.~Xiong, and Q.~Dai, ``Residual highway convolutional neural networks for in-loop filtering in hevc,'' \emph{IEEE Transactions on image processing}, vol.~27, no.~8, pp. 3827--3841, 2018.

\bibitem{ma2020mfrnet}
D.~Ma, F.~Zhang, and D.~R. Bull, ``Mfrnet: a new cnn architecture for post-processing and in-loop filtering,'' \emph{IEEE Journal of Selected Topics in Signal Processing}, vol.~15, no.~2, pp. 378--387, 2020.

\bibitem{li2021deepqtmt}
T.~Li, M.~Xu, R.~Tang, Y.~Chen, and Q.~Xing, ``Deepqtmt: A deep learning approach for fast qtmt-based cu partition of intra-mode vvc,'' \emph{IEEE Transactions on Image Processing}, vol.~30, pp. 5377--5390, 2021.

\bibitem{jia2019content}
C.~Jia, S.~Wang, X.~Zhang, S.~Wang, J.~Liu, S.~Pu, and S.~Ma, ``Content-aware convolutional neural network for in-loop filtering in high efficiency video coding,'' \emph{IEEE Transactions on Image Processing}, vol.~28, no.~7, pp. 3343--3356, 2019.

\bibitem{ma2019convolutional}
C.~Ma, D.~Liu, X.~Peng, L.~Li, and F.~Wu, ``Convolutional neural network-based arithmetic coding for hevc intra-predicted residues,'' \emph{IEEE Transactions on Circuits and Systems for Video Technology}, vol.~30, no.~7, pp. 1901--1916, 2019.

\bibitem{liu2025frequency}
J.~Liu, Q.~Zheng, Z.~Liu, Y.~Zhong, P.~Liu, T.~Liu, S.~Xu, Y.~Lu, S.~Li, D.~Niu \emph{et~al.}, ``Frequency-biased synergistic design for image compression and compensation,'' in \emph{Proceedings of the Computer Vision and Pattern Recognition Conference}, 2025, pp. 12\,820--12\,829.

\bibitem{yang2023lossy}
R.~Yang and S.~Mandt, ``Lossy image compression with conditional diffusion models,'' \emph{Advances in Neural Information Processing Systems}, vol.~36, pp. 64\,971--64\,995, 2023.

\bibitem{wan2025m3}
R.~Wan, Q.~Zheng, and Y.~Fan, ``M3-cvc: Controllable video compression with multimodal generative models,'' in \emph{ICASSP 2025-2025 IEEE International Conference on Acoustics, Speech and Signal Processing (ICASSP)}.\hskip 1em plus 0.5em minus 0.4em\relax IEEE, 2025, pp. 1--5.

\bibitem{zheng2025unicorn}
Q.~Zheng, H.~Wang, Z.~Liu, J.~Liu, Z.~Hao, B.~Chen, M.~Li, R.~Wan, P.~Liu, Y.~Lu \emph{et~al.}, ``Unicorn: Unified neural image compression with one number reconstruction,'' in \emph{Proceedings of the 33rd ACM International Conference on Multimedia}, 2025, pp. 8626--8635.

\bibitem{kingma2013auto}
D.~P. Kingma and M.~Welling, ``Auto-encoding variational bayes,'' \emph{arXiv preprint arXiv:1312.6114}, 2013.

\bibitem{wallace1991jpeg}
G.~K. Wallace, ``The jpeg still picture compression standard,'' \emph{Communications of the ACM}, vol.~34, no.~4, pp. 30--44, 1991.

\bibitem{skodras2001jpeg2000}
A.~N. Skodras, C.~A. Christopoulos, and T.~Ebrahimi, ``Jpeg2000: The upcoming still image compression standard,'' \emph{Pattern recognition letters}, vol.~22, no.~12, pp. 1337--1345, 2001.

\bibitem{bellard2015bpg}
F.~Bellard, ``Bpg image format,'' \emph{URL https://bellard. org/bpg}, vol.~1, no.~2, p.~1, 2015.

\bibitem{hu2021fvc}
Z.~Hu, G.~Lu, and D.~Xu, ``Fvc: A new framework towards deep video compression in feature space,'' in \emph{Proceedings of the IEEE/CVF Conference on Computer Vision and Pattern Recognition}, 2021, pp. 1502--1511.

\bibitem{liu2021deep}
B.~Liu, Y.~Chen, S.~Liu, and H.-S. Kim, ``Deep learning in latent space for video prediction and compression,'' in \emph{Proceedings of the IEEE/CVF conference on computer vision and pattern recognition}, 2021, pp. 701--710.

\bibitem{goodfellow2014generative}
I.~Goodfellow, J.~Pouget-Abadie, M.~Mirza, B.~Xu, D.~Warde-Farley, S.~Ozair, A.~Courville, and Y.~Bengio, ``Generative adversarial nets,'' \emph{Advances in neural information processing systems}, vol.~27, 2014.

\bibitem{siarohin2019first}
A.~Siarohin, S.~Lathuili{\`e}re, S.~Tulyakov, E.~Ricci, and N.~Sebe, ``First order motion model for image animation,'' \emph{Advances in neural information processing systems}, vol.~32, 2019.

\bibitem{konuko2022hybrid}
G.~Konuko, S.~Lathuili{\`e}re, and G.~Valenzise, ``A hybrid deep animation codec for low-bitrate video conferencing,'' \emph{arXiv preprint arXiv:2207.13530}, 2022.

\bibitem{konuko2023predictive}
------, ``Predictive coding for animation-based video compression,'' in \emph{2023 IEEE International Conference on Image Processing (ICIP)}.\hskip 1em plus 0.5em minus 0.4em\relax IEEE, 2023, pp. 2810--2814.

\bibitem{chen2022beyond}
B.~Chen, Z.~Wang, B.~Li, R.~Lin, S.~Wang, and Y.~Ye, ``Beyond keypoint coding: Temporal evolution inference with compact feature representation for talking face video compression,'' in \emph{2022 Data Compression Conference (DCC)}.\hskip 1em plus 0.5em minus 0.4em\relax IEEE, 2022, pp. 13--22.

\bibitem{chen2023compact}
B.~Chen, Z.~Wang, B.~Li, S.~Wang, and Y.~Ye, ``Compact temporal trajectory representation for talking face video compression,'' \emph{IEEE Transactions on Circuits and Systems for Video Technology}, 2023.

\bibitem{wang2023extreme}
R.~Wang, Q.~Mao, C.~Jia, R.~Wang, and S.~Ma, ``Extreme generative human-oriented video coding via motion representation compression,'' in \emph{2023 IEEE International Symposium on Circuits and Systems (ISCAS)}.\hskip 1em plus 0.5em minus 0.4em\relax IEEE, 2023, pp. 1--5.

\bibitem{volokitin2022neural}
A.~Volokitin, S.~Brugger, A.~Benlalah, S.~Martin, B.~Amberg, and M.~Tschannen, ``Neural face video compression using multiple views,'' in \emph{Proceedings of the IEEE/CVF Conference on Computer Vision and Pattern Recognition}, 2022, pp. 1738--1742.

\bibitem{wang2022dynamic}
Z.~Wang, B.~Chen, Y.~Ye, and S.~Wang, ``Dynamic multi-reference generative prediction for face video compression,'' in \emph{2022 IEEE International Conference on Image Processing (ICIP)}.\hskip 1em plus 0.5em minus 0.4em\relax IEEE, 2022, pp. 896--900.

\bibitem{tang2022generative}
A.~Tang, Y.~Huang, J.~Ling, Z.~Zhang, Y.~Zhang, R.~Xie, and L.~Song, ``Generative compression for face video: A hybrid scheme,'' in \emph{2022 IEEE International Conference on Multimedia and Expo (ICME)}.\hskip 1em plus 0.5em minus 0.4em\relax IEEE, 2022, pp. 1--6.

\bibitem{oquab2022efficient}
M.~Oquab, D.~Haziza, L.~Schwartz, T.~Xu, K.~Zand, R.~Wang, P.~Liu, and C.~Couprie, ``Efficient conditioned face animation using frontally-viewed embedding,'' \emph{arXiv preprint arXiv:2203.08765}, 2022.

\bibitem{jia2022fpx}
C.~Jia, X.~Hang, S.~Wang, Y.~Wu, S.~Ma, and W.~Gao, ``Fpx-nic: An fpga-accelerated 4k ultra-high-definition neural video coding system,'' \emph{IEEE Transactions on Circuits and Systems for Video Technology}, vol.~32, no.~9, pp. 6385--6399, 2022.

\bibitem{electronics12102289}
\BIBentryALTinterwordspacing
H.~Shao, B.~Liu, Z.~Li, C.~Yan, Y.~Sun, and T.~Wang, ``A high-throughput processor for gdn-based deep learning image compression,'' \emph{Electronics}, vol.~12, no.~10, 2023. [Online]. Available: \url{https://www.mdpi.com/2079-9292/12/10/2289}
\BIBentrySTDinterwordspacing

\bibitem{steven2016convolutional}
K.~Steven, P.~Merolla, J.~Arthur, and A.~Cassidy, ``Convolutional networks for fast, energy-efficient neuromorphic computing,'' \emph{Proc. Natl. Acad. Sci. USA}, vol.~27, p. 201604850, 2016.

\bibitem{263_1}
\BIBentryALTinterwordspacing
A.~{Ben Atitallah}, P.~Kadionik, F.~Ghozzi, P.~Nouel, N.~Masmoudi, and H.~Levi, ``An fpga implementation of hw/sw codesign architecture for h.263 video coding,'' \emph{AEU - International Journal of Electronics and Communications}, vol.~61, no.~9, pp. 605--620, 2007. [Online]. Available: \url{https://www.sciencedirect.com/science/article/pii/S1434841106001191}
\BIBentrySTDinterwordspacing

\bibitem{264_1}
T.-C. Chen, C.-J. Lian, and L.-G. Chen, ``Hardware architecture design of an h. 264/avc video codec,'' in \emph{Proceedings of the 2006 Asia and South Pacific Design Automation Conference}, 2006, pp. 750--757.

\bibitem{hevc1}
\BIBentryALTinterwordspacing
A.~Ben~Atitallah, M.~Kammoun, K.~M. Ali, and R.~Ben~Atitallah, ``An fpga comparative study of high-level and low-level combined designs for hevc intra, inverse quantization, and idct/idst 2d modules,'' \emph{International Journal of Circuit Theory and Applications}, vol.~48, no.~8, pp. 1274--1290, 2020. [Online]. Available: \url{https://onlinelibrary.wiley.com/doi/abs/10.1002/cta.2790}
\BIBentrySTDinterwordspacing

\bibitem{hevc2}
A.~B. Atitallah and M.~Kammoun, ``High-level design of hevc intra prediction algorithm,'' in \emph{2020 5th International Conference on Advanced Technologies for Signal and Image Processing (ATSIP)}, 2020, pp. 1--6.

\bibitem{vvc1}
H.~Azgin, E.~Kalali, and I.~Hamzaoglu, ``An approximate versatile video coding fractional interpolation hardware,'' in \emph{2020 IEEE International Conference on Consumer Electronics (ICCE)}.\hskip 1em plus 0.5em minus 0.4em\relax IEEE, 2020, pp. 1--4.

\bibitem{vvc2}
Y.~Fan, Y.~Zeng, H.~Sun, J.~Katto, and X.~Zeng, ``A pipelined 2d transform architecture supporting mixed block sizes for the vvc standard,'' \emph{IEEE Transactions on Circuits and Systems for Video Technology}, vol.~30, no.~9, pp. 3289--3295, 2019.

\bibitem{vvc3}
I.~Farhat, W.~Hamidouche, A.~Grill, D.~Menard, and O.~D{\'e}forges, ``Lightweight hardware implementation of vvc transform block for asic decoder,'' in \emph{ICASSP 2020-2020 IEEE International Conference on Acoustics, Speech and Signal Processing (ICASSP)}.\hskip 1em plus 0.5em minus 0.4em\relax IEEE, 2020, pp. 1663--1667.

\bibitem{vvc4}
H.~Azgin, E.~Kalali, and I.~Hamzaoglu, ``An efficient fpga implementation of versatile video coding intra prediction,'' in \emph{2019 22nd Euromicro Conference on Digital System Design (DSD)}.\hskip 1em plus 0.5em minus 0.4em\relax IEEE, 2019, pp. 194--199.

\bibitem{F-CAD}
X.~Zhang, D.~Wang, P.~Chuang, S.~Ma, D.~Chen, and Y.~Li, ``F-cad: A framework to explore hardware accelerators for codec avatar decoding,'' in \emph{2021 58th ACM/IEEE Design Automation Conference (DAC)}, 2021, pp. 763--768.

\bibitem{siarohin2019animating}
A.~Siarohin, S.~Lathuili{\`e}re, S.~Tulyakov, E.~Ricci, and N.~Sebe, ``Animating arbitrary objects via deep motion transfer,'' in \emph{Proceedings of the IEEE/CVF Conference on Computer Vision and Pattern Recognition}, 2019, pp. 2377--2386.

\bibitem{ronneberger2015u}
O.~Ronneberger, P.~Fischer, and T.~Brox, ``U-net: Convolutional networks for biomedical image segmentation,'' in \emph{Medical Image Computing and Computer-Assisted Intervention--MICCAI 2015: 18th International Conference, Munich, Germany, October 5-9, 2015, Proceedings, Part III 18}.\hskip 1em plus 0.5em minus 0.4em\relax Springer, 2015, pp. 234--241.

\bibitem{johnson2016perceptual}
J.~Johnson, A.~Alahi, and L.~Fei-Fei, ``Perceptual losses for real-time style transfer and super-resolution,'' in \emph{Computer Vision--ECCV 2016: 14th European Conference, Amsterdam, The Netherlands, October 11-14, 2016, Proceedings, Part II 14}.\hskip 1em plus 0.5em minus 0.4em\relax Springer, 2016, pp. 694--711.

\bibitem{mao2017least}
X.~Mao, Q.~Li, H.~Xie, R.~Y. Lau, Z.~Wang, and S.~Paul~Smolley, ``Least squares generative adversarial networks,'' in \emph{Proceedings of the IEEE international conference on computer vision}, 2017, pp. 2794--2802.

\bibitem{markuvs2018fusing}
N.~Marku{\v{s}}, ``Fusing batch normalization and convolution in runtime,'' 2018.

\bibitem{repair}
K.~Jordan, H.~Sedghi, O.~Saukh, R.~Entezari, and B.~Neyshabur, ``Repair: Renormalizing permuted activations for interpolation repair,'' \emph{arXiv preprint arXiv:2211.08403}, 2022.

\bibitem{Nagrani_2017}
\BIBentryALTinterwordspacing
A.~Nagrani, J.~S. Chung, and A.~Zisserman, ``Voxceleb: A large-scale speaker identification dataset,'' in \emph{Interspeech 2017}.\hskip 1em plus 0.5em minus 0.4em\relax ISCA, Aug. 2017. [Online]. Available: \url{http://dx.doi.org/10.21437/Interspeech.2017-950}
\BIBentrySTDinterwordspacing

\bibitem{wang2003multiscale}
Z.~Wang, E.~P. Simoncelli, and A.~C. Bovik, ``Multiscale structural similarity for image quality assessment,'' in \emph{The Thrity-Seventh Asilomar Conference on Signals, Systems \& Computers, 2003}, vol.~2.\hskip 1em plus 0.5em minus 0.4em\relax Ieee, 2003, pp. 1398--1402.

\bibitem{zhang2018unreasonable}
R.~Zhang, P.~Isola, A.~A. Efros, E.~Shechtman, and O.~Wang, ``The unreasonable effectiveness of deep features as a perceptual metric,'' in \emph{Proceedings of the IEEE conference on computer vision and pattern recognition}, 2018, pp. 586--595.

\bibitem{blau2018perception}
Y.~Blau and T.~Michaeli, ``The perception-distortion tradeoff,'' in \emph{Proceedings of the IEEE conference on computer vision and pattern recognition}, 2018, pp. 6228--6237.

\bibitem{blau2019rethinking}
------, ``Rethinking lossy compression: The rate-distortion-perception tradeoff,'' in \emph{International Conference on Machine Learning}.\hskip 1em plus 0.5em minus 0.4em\relax PMLR, 2019, pp. 675--685.

\bibitem{zhang2021universal}
G.~Zhang, J.~Qian, J.~Chen, and A.~Khisti, ``Universal rate-distortion-perception representations for lossy compression,'' \emph{Advances in Neural Information Processing Systems}, vol.~34, pp. 11\,517--11\,529, 2021.

\end{thebibliography}


\end{document}